\documentclass[journal]{IEEEtran}

\ifCLASSINFOpdf

\else

\fi

\usepackage{xcolor}

\usepackage{amssymb,amsthm}
\usepackage{diagbox}

\newtheorem{definition}{Definition}

\newtheorem{remark}{Remark}
\usepackage{algorithm}
\usepackage{algorithmic}
\usepackage{ulem}
\usepackage{multirow}

\usepackage[utf8]{inputenc}
\usepackage[T1]{fontenc}    
\usepackage{url}            
\usepackage{booktabs}       
\usepackage{amsfonts}       
\usepackage{nicefrac}      
\usepackage{microtype}      

\usepackage{times}
\usepackage{soul}
\usepackage{fdsymbol}
\usepackage{graphicx}
\usepackage{amsmath}

\usepackage{import}
\usepackage{subfigure}

\labelformat{equation}{(#1)}

\begin{document}
\title{THP: Topological Hawkes Processes for Learning Causal Structure on Event Sequences}

\author{
        Ruichu Cai,~\IEEEmembership{Senior Member,~IEEE},
        Siyu Wu,
		Jie Qiao,
        Zhifeng Hao~\IEEEmembership{Senior Member,~IEEE},
        Keli Zhang,
        Xi Zhang
\thanks{Ruichu Cai and Jie Qiao are with the School of Computer Science, Guangdong University of Technology, Guangzhou, China, 510006 and Peng Cheng Laboratory, Shenzhen, China, 518066. \protect E-mail: \{cairuichu, qiaojie.chn\}@gmail.com.}
\thanks{Siyu Wu is with the School of Computer Science, Guangdong University of Technology, Guangzhou, China, 510006. \protect E-mail: fisherwsy@163.com.}
\thanks{Zhifeng Hao is with College of Science, Shantou University, Shantou, China, 515063. \protect E-mail: haozhifeng@stu.edu.cn.}
\thanks{Keli Zhang and Xi Zhang are with Huawei Noah's Ark Lab, Huawei, Shenzhen, China, 518116. \protect  E-mail: \{zhangkeli1, zhangxi28\}@huawei.com.}
\thanks{This research was supported in part by National Key R\&D Program of China (2021ZD0111501), National Science Fund for Excellent Young Scholars (62122022), Natural Science Foundation of China (61876043, 61976052), the major key project of PCL (PCL2021A12).}
}

\markboth{Journal of \LaTeX\ Class Files,~Vol.~14, No.~8, August~2015}
{Shell \MakeLowercase{\textit{et al.}}: Bare Demo of IEEEtran.cls for IEEE Journals}

\maketitle

\begin{abstract}
 Learning causal structure among event types on multi-type event sequences is an important but challenging task. Existing methods, such as the Multivariate Hawkes processes, mostly assumed that each sequence is independent and identically distributed. However, in many real-world applications, it is commonplace to encounter a topological network behind the event sequences such that an event is excited or inhibited not only by its history but also by its topological neighbors. Consequently, the failure in describing the topological dependency among the event sequences leads to the error detection of the causal structure. By considering the Hawkes processes from the view of temporal convolution, we propose a Topological Hawkes process (THP) to draw a connection between the graph convolution in the topology domain and the temporal convolution in time domains. We further propose a causal structure learning method on THP in a likelihood framework. The proposed method is featured with the graph convolution-based likelihood function of THP and a sparse optimization scheme with an Expectation-Maximization of the likelihood function. Theoretical analysis and experiments on both synthetic and real-world data demonstrate the effectiveness of the proposed method.
\end{abstract}


\IEEEpeerreviewmaketitle

\section{Introduction}
\IEEEPARstart{L}{earning} causal structure among event types on \textit{multi-type event sequences} is an important task in many real-world applications. For example, social scientists may be interested in the study of the causal relationships among social events \cite{zhou2013learningsocial}, economists may be interested in analyzing the causality among economic time series \cite{bacry2015hawkes}, and the network operation and maintenance engineers try to locate the root cause of the alarm events that occur intensively. \cite{chang2010causal}. 

Various methods have been proposed to discover the causality on multi-type event sequences. One line of research is the constraint-based methods, focusing on exploring the independence within the causal variables. The typical methods include PCMCI (PC \& Momentary Conditional Independence test) \cite{runge2019detecting}, and the transfer entropy based methods \cite{novelli2019large,mijatovic2021information,shorten2021estimating}. Another line is the point processes based methods, focusing on modeling the generation process of the events. The typical methods include the Multivariate Hawkes processes based methods \cite{xu2016learning,bhattacharjya2018proximal,salehi2019learning}. Our work belongs to the Multivariate Hawkes processes based methods.

\begin{figure*}[tb]
	\centering
	\includegraphics[width=\textwidth]{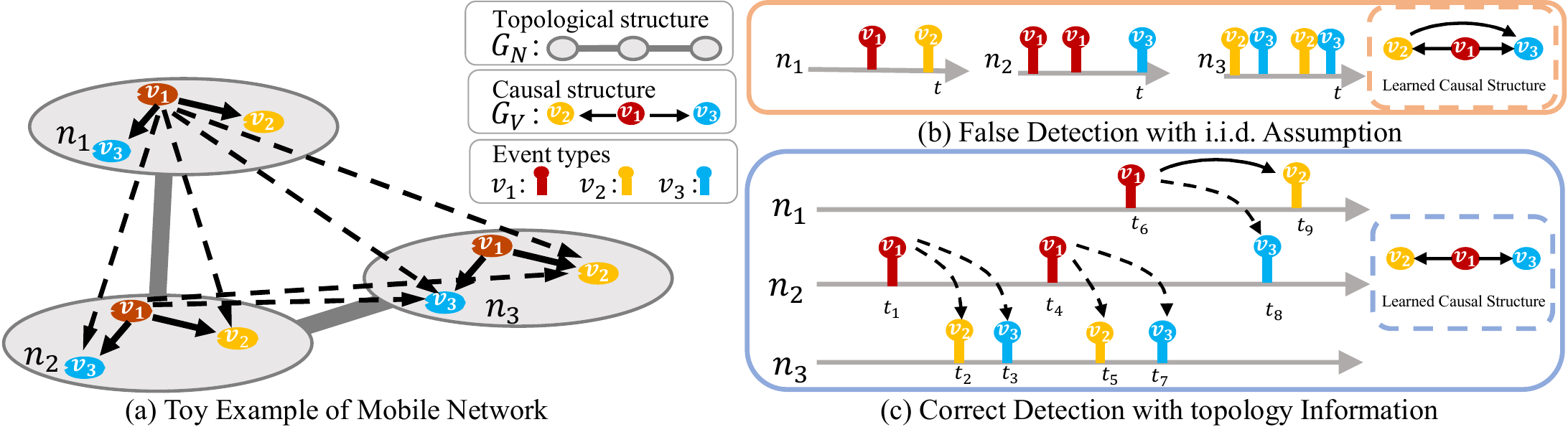}
	\caption{A toy example of the topological-temporal event data. In the example, the gray circles are the network elements connected by the topological graph $\mathcal{G}_N$ (e.g., the mobile networks), and the dots are the events (e.g. the different types of alarms) influenced by the causal structure $\mathcal{G}_V$ behind the events. The solid arrows indicate the impact of the alarms on directly connected nodes, while the dashed 
arrows indicate the impact on its topological neighbors.}
	\label{fig:fig1}      
\end{figure*}

One common assumption of these existing methods is that the event sequences are independent and identically distributed (i.i.d.). However, in many real-world scenarios, the event sequences are usually generated by nodes in a topological network such that an event will not only be excited or inhibited by the events inside the sequence but also by the events in its topological neighbors.

For example, considering the mobile network given in Fig. \ref{fig:fig1}(a), there are three types of alarms $v_1,v_2,v_3$ spreading across the network stations $\mathcal{G}_N$ with the causal structure $\mathcal{G}_V$. One would like to identify the causal relationships among alarm types to help operate and maintain the mobile network. Here, each node represents a network element in the mobile network and the edge among the nodes represents a physical connection. Moreover, in such a case, the alarm events sequences produced by nodes $n_1,n_2,n_3$ are no longer independent as a cascading failure would occur from one node to another node through the physical connection between each network element. That is, the causal relationship of the failure alarm types $v_1\to v_2$ will also appear among the neighbors (the dashed line). Then, if we ignore the topological structure behind the sequences and treat the sequences of different nodes independently, the existing methods may introduce unobservable confounders and return vulnerable and unstable results.
As illustrated in Fig. \ref{fig:fig1}(b), if we ignore the topology information and falsely treat the sequences independently, it turns out that a falsely discovered causal relationship $v_2\to v_3$ will be made. Specifically, under the i.i.d. assumption, for $v_2$ and $v_3$ in $n_3$, $v_1$ in $n_2$ turns to be an unobservable confounder leading to the false detection of strong dependence between $v_2$ and $v_3$. Such a phenomenon can also be found in many scenarios. For example, in social science, the behavior of individuals may influence each other in some way through social networks (topological 
structures of interpersonal relationships).

Thus, it is crucial to consider the topological structure behind the data for learning causal structure. However, modeling the topological information is a non-trivial task, due to the complex propagation of the causal influence.

To fill this gap, we extend the time-domain convolution property of the Hawkes process \cite{xu2016learning} to the time-graph domain to tackle the topological dependency and finally propose the Topological Hawkes processes (THP).
In such a process, the events' generation process is constrained by the topological structure $\mathcal{G}_N$ and the causal structure $\mathcal{G}_V$, as illustrated in Fig. \ref{fig:fig1}(c). We further derive the likelihood function of THP using both the time convolution and the graph convolution. Such a likelihood model is optimized by an Expectation-Maximization based optimization scheme. 

As a summary, our contributions of this paper include, 1) proposing a Topological Hawkes Processes (THP) for the event sequences generated by the topological networks, 2) deriving the likelihood function of THP using the joint convolution on the graph-time domain, 3) developing an effective sparse optimization schema for the above likelihood function, and  4) conducting theoretical analysis and extensive experiments on the proposed model.

\section{Related Work}
\paragraph{Point  processes} There are mainly two types of point processes for the complicated event sequences in recent researches. The first type is the Hawkes processes \cite{hawkes1971spectra} which assumes that the historical events have influences on the future ones. It uses a parametric or non-parametric intensity function to model the self-exciting and mutually-exciting mechanism of the point process in 1-dimensional \cite{lewis2011nonparametric,zhao2015seismic,zhang2019efficient} and multi-dimensional case \cite{farajtabar2014shaping,zhou2013learning,luo2015multi}. The second type of point processes is the deep-learning-based point process methods \cite{du2016recurrent,mei2017neural,shang2019geometric}, which use a learnable intensity function to capture the nonlinear impact from past to future. While GHP (Geometric Hawkes Processes) \cite{shang2019geometric} uses the graph convolution network to model the relation among the parameters of self-exciting Hawkes processes, it does not model the dependency among the multivariate processes.

\paragraph{Granger causality} To learn the Granger causality between event type, multivariate Hawkes processes are proposed which is based on the Hawkes process \cite{hawkes1971spectra}. The variant of these methods mainly focus on the design of regularization and the intensity function. For example, ADM4 (Alternating Direction Method of Multipliers and Majorization Minimization) \cite{zhou2013learningsocial} use a nuclear and $l1$ norm as the sparse regularization term, MLE-SGL (Maximum Likelihood Estimator with Sparse-Group-Lasso) \cite{xu2016learning} further considers the temporal sparsity and pairwise similarity regularization, and NPHC (Non-Parametric Hawkes Cumulant) \cite{achab2017uncovering} proposes a new non-parametric method to estimate the matrix of integrated kernels of multivariate Hawkes process. Recently, some deep point process based methods have also been proposed. For example, RPPN (Recurrent Point Process Networks) \cite{xiao2019learning} introduces the attention mechanism to preserve the quasi-causality of events, and CAUSE (Causality from AttribUtions on Sequence of Events) \cite{02817a8a3ee244c68c2a91510915c943} uses the attribution method combined with a deep neural network to learn the Granger causality. Our work is also related to the PGEM (Proximal Graphical Event Model) \cite{bhattacharjya2018proximal} which uses a greedy algorithm to search the causal graph but assumes the event types depend only on the most recent history with fixed window size. Some works also consider learning point process under the non-i.i.d. samples, e.g., \cite{bacry2016first} introduces the first- and second-order properties to characterize a Hawkes process, while \cite{etesami2016learning} proposes to learn a directed information graph under heterogeneous data, but neither of them considers the topological graph behind the data. In addition to the point process based methods, some methods are proposed to learn the Granger causality from event sequences nonparametrically by using the transfer entropy \cite{mijatovic2021information,shorten2021estimating}.

\section{Preliminary}

\subsection{Multivariate Point Processes}\label{sec:mpp}
A multivariate point process is a random process whose realization can be presented by $\mathcal{E} = \{v_i,t_i\}^{m}_{i=1}$, where $v_i \in \mathbf{V}$ and  $t_i\in \mathbf{T}$ are the type and the occurrence time of the $i$-th event, $m$ is the number of events. It can be equivalently represented by a $|\mathbf{V}|$-dimensional counting process $C=\{C_v(t)|t\in \mathbf{T},v\in \mathbf{V} \}$ where $C_v(t)\in \mathbb{N}$ records the number of events 
that occur before time $t$ in the event type $v$. One way to model the above counting process is to use the following \textit{conditional intensity function} to characterize the probability of the count of event occurrence given the history:
\begin{equation*}\label{eq: lambda}
\small
  \lambda_{v}(t)dt = \lambda_{v}(t|\mathcal{H}_{t}^{\mathbf{V}})dt =\mathbb{E}[dC_{v}(t)|{\mathcal{F}(t)}],  
\end{equation*}
where $\mathcal{H}_{t}^{\mathbf{V}}=\{t_i,v_i|t_i<t,v_i\in \mathbf{V}\}$ collects the historical events of all event types before time $t$, $dC_v(t)=C_v(t+dt)-C_v(t)$, and $\mathcal{F}=\{\mathcal{F}^1,...,\mathcal{F}^{|\mathbf{V}|}\}$ is the collection of canonical filtration of all counting processes.
We further denote $\mathcal{F}(t)$ as the canonical filtration of sub-processes $C_v(t)$ up to time $t$ such that $\mathcal{F}(t_1)\subset \mathcal{F}(t_2)$ for every $t_1<t_2$.

A typical class of the multivariate point processes is Hawkes processes. It considers a particular type of intensity function which measures the excitation from the past event:
\begin{align}
\small
\label{eq:intensity_continue}
\lambda _{v} (t) & =\mu _{v} +\sum _{v'\in \mathbf{V}}\int _{t'\in \mathbf{T}_{t-}} \phi _{v',v} (t-t')dC_{v'}( t'),
\end{align}

where $\mathbf{T}_{t^-}=\{t'|t' \in \mathbf{T},t'<t\}$, and the intensity $\lambda _{v}(t)$ is a summation over the base intensity $\mu_{v}$ and the endogenous intensity $\sum _{v'\in \mathbf{V}}\int _{t'\in \mathbf{T}_{t^-}} \phi _{v',v} (t-t')dC_{v'}( t')$ aiming to capture the peer influence occurring near time $t$ \cite{farajtabar2014shaping}.
$\phi_{v',v}(t)$ is an impact function characterizing the time-decay of the causal influence. 

\subsection{Graph Convolution}

Formally, we use an undirected graph $\mathcal{G}_N =(\mathbf{N} ,\mathbf{E}_{N} )$ to represent the topological structure over a set of nodes $\mathbf{N}$ and edges $\mathbf{E}_{N}$. Let $A$ denote the adjacency matrix of $\mathcal{G}_N$ in which $A_{ij}=1$ if $(i,j)\in \mathbf{E}_{N}$. The normalized Laplacian matrix can be defined as $L=I-D^{-1/2}AD^{-1/2}$, where $D\in \mathbb{R}^{|\mathbf{N}|\times|\mathbf{N}|}$ is the diagonal degree matrix with $D_{ii}=\sum_jA_{ij}$ and $I$ is the identity matrix. As $L$ is a real symmetric positive semidefinite matrix, it has a set of orthonormal eigenvectors $\{u_l\}_{l=0}^{|\mathbf{N}|-1}$ with the orthonormal eigenvectors matrix $U\in \mathbb{R}^{|\mathbf{N}|\times|\mathbf{N}|}$, and the associated eigenvalues $\{\gamma_l\}_{l=0}^{|\mathbf{N}|-1}$ with the diagonalization eigenvalues matrix $\Gamma =\text{Diag}([\gamma_0,...,\gamma_{|\mathbf{N}|-1}])$.

The graph convolution is built upon the spectral convolutions using the graph Fourier transform \cite{sandryhaila2013discrete,defferrard2016convolutional} such that the graph Fourier transform of a signal $s\in \mathbb{R}^{|N|}$ can be defined as $\hat{s}=U^Ts$ and its inverse as $s=U\hat{s}$. Then, graph convolution over $\mathcal{G}_N$ is $(s_1*s_2)_{\mathcal{G}_N}=U((U^Ts_1)\odot (U^Ts_2))$, where $\odot$ is the element-wise Hadamard product and $s_1,s_2$ are the signal vectors. It follows that a signal $s$ is filtered by $g_\theta$ as:

\begin{equation}
\small
\label{eq:graph convolution}
    y=g_{\theta}(L)s=g_{\theta}(U\Gamma U^T)s=U g_{\theta}(\Gamma)U^Ts,
\end{equation}
where $g_{\theta}$ is the graph convolution kernel. Note that $g_{\theta}$ is a function defined on graph domain. Intuitively, an graph convolution can be interpreted as a filter of the Laplace matrix such that the signal on each node is propagated through the graph.

\section{Topological Hawkes Process}

\subsection{Problem Formalization}
In this work, we consider the real-world scenarios that the event sequences are usually generated by nodes in a topological network such that an event will not only be excited or inhibited by the cause event that happens in the sequence itself but also by the cause event that is in its topological neighbors. To formalize the above scenarios, we use an undirected graph $\mathcal{G}_N =(\mathbf{N} ,\mathbf{E}_{N} )$ to represent the topological graph among the nodes $\mathbf{N}$ and a directed graph $\mathcal{G}_V =(\mathbf{V} ,\mathbf{E}_{V})$ to represent the causal structure among event types $\mathbf{V}$, where $\mathbf{E}_{N}$ and $\mathbf{E}_V$ are the sets of edges in the topological graph and the causal graph respectively. Based on the above notations, we further extend the traditional event sequences $\mathcal{E} = \{v_i,t_i\}^{m}_{i=1}$ to $\mathcal{E} = \{n_i,v_i,t_i\}^{m}_{i=1}$, by using  $n_i \in \mathbf{N}$ to denote the corresponding node in the topological graph. Then, we formalize our problem as follows:

\begin{definition}[Learning causal structure on topological event sequences]
Given a set of observed event sequences $\mathcal{E}=\{n_i,v_i,t_i\}^{m}_{i=1},n_i \in  \mathbf{N}, v_i \in  \mathbf{V}, t_i \in  \mathbf{T}$ and their corresponding topological graph $\mathcal{G}_N$, the goal of this work is to discover the causal structure $\mathcal{G}_V$ among the event types $\mathbf{V}$.
\end{definition}

\subsection{Formalization of Topological Hawkes Process}
It is easy to see that the existing multivariate Hawkes processes are limited for the above problem. Because the existing multivariate Hawkes processes assume the set of sequences are i.i.d. and ignore the underlying topological structure behind the data generating process, which may lead to false detection as mentioned in Fig. \ref{fig:fig1}(b). Thus, by introducing the topological graph into the Hawkes process, we propose the topological Hawkes processes to handle the topological event sequences. However, introducing the topological graph into the Hawkes processes is a non-trivial task, as an event could be excited by its cause event from its topological neighbor through different paths. Such observation implies that the intensity of an event type can be viewed as the summation of the cause event type intensity over different paths, which inspires a way to model it using the graph convolution.

To introduce the graph convolution into the multivariate Hawkes process, we begin with the temporal convolution view of the traditional Hawkes process. From the perspective of temporal convolution, the intensity function in Eq. \ref{eq:intensity_continue} can be equivalently formalized as follows:
\begin{equation}
\small
\label{eq:intensity_convolution}
    \lambda _{v} (t)=\mu _{v} +\sum _{v'\in \mathbf{V}} (\phi _{v',v} *dC_{v'} )_{\mathbf{T}} (t).
\end{equation}
The above formalization reveals that the intensity function of Hawkes processes is essentially a convolution operation in time domain, which means that the intensity function is the cumulative effect from all the past causally related events at time $t$. Here $(\phi _{v',v} *dC_{v'} )_{\mathbf{T}}(t)$ denotes a function of $t$ which is obtained from the convolution of function $\phi _{v',v}$ and $dC_{v'}$. By extending the convolution in time domain to the join convolution in the graph-time domain $\mathcal{G}_N\times \mathbf{T}$, the intensity of the event type $v$ at node $n$ at time $t$ is derived as follows: 
\begin{equation}
\small
\label{eq:join_convolution}
\lambda _{v} (n,t) =\mu _{v} +\sum _{v'\in \mathbf{V}}(\psi_{v',v}*dC_{v'})_{\mathcal{G}_N\times \mathbf{T}}(n,t),
\end{equation}
where $dC_{v'}$ is a function defined on $\mathcal{G}_N\times \mathbf{T}$ such that $dC_{v'}(n,t)$ denotes the count of occurrence events of type $v$ event in node $n$ in time interval $[t-dt, t]$, $\psi_{v',v}$ is the convolution kernel which measures the causal impact from $v'$ in both the past and neighbors. Similiarly, $(\psi_{v',v}*dC_{v'})_\mathbf{\mathcal{G}_N\times \mathbf{T}}(n,t)$ depicts a function of $n,t$ obtained from the convolution of $\phi _{v',v}$ and $dC_{v'}$ in both graph and time domain.

Practically, by assuming the invariance of the topological graph across the time, such joint convolution operation can be decomposed into two folds:
\begin{gather}
  \label{eq:convolution_time}
s_{v',v,t}=( \phi _{v',v} *dC_{v'})_{\mathbf{T}}( t),\\
  \lambda _{v} (n,t)=\mu _{v} +\sum _{v'\in \mathbf{V}}( g_{v',v} *s_{v',v,t})_{\mathcal{G}_{N}} (n),
\label{eq:convolution_graph}
\end{gather}
where $g_{v',v}$ and $\phi_{v',v}$ are the convolution kernels on the graph domain $\mathcal{G}_N$ and the time domain $\mathbf{T}$, respectively. First, in Eq. \ref{eq:convolution_time}, $s_{v',v,t}$ can be viewed as the summation of the impact from the past with the impact function $\phi_{v',v}(t)$, which is the vanilla endogenous intensity of multivariate Hawkes processes (see Sec. \ref{sec:mpp}). Second, $s_{v',v,t}$ will be considered as the signal function in graph convolution. Then, after considering all the impact from causal event types, we have the overall intensity $\lambda_v(n,t)$ in Eq. \ref{eq:convolution_graph}. Intuitively, the convolution kernel $\phi_{v',v}$ in the 
time domain is used to summarize the effects of past events to obtain $s_{v',v,t}$, and the convolution kernel $g_{v',v}$ in the graph domain is used to aggregate the effects of topological neighbors denoted by $s_{v',v,t}$. Note that different convolution kernels correspond to different forms of aggregation that
can be used to portray different impact mechanisms.

An important question is how to design the convolution kernel. As for the \textit{temporal convolution kernel}, referring to other works, we set it to $\phi_{v',v}(t)=\begin{cases} \beta_{v',v}\kappa(t) & t >0\\ 0 & t\leqslant 0 \end{cases}$, where $\beta_{v',v}$ is the causal strength of edge $v'\to v$ and $\kappa(t)$ characterize the time-decay of the causal influence usually using the exponential form $\kappa (t)=\exp (\delta t) $ with the hyper-parameter $\delta$. As for the \textit{graph convolution kernel}, recall Eq. \ref{eq:graph convolution}, the graph convolution in Eq. \ref{eq:convolution_graph} can be rewritten as $
\lambda _{v} (n,t)=\mu _{v} +\sum _{v'\in \mathbf{V}}(g_{v',v}( L ) s_{v',v,t})(n)$, which represents the accumulated effect from the past events in time $t$ at node $n$.
Here, the graph kernel $g_{v',v}$ characterizes the signal propagation within the topological graph, which is flexible and easy to be extended, e.g., using the graph convolutional networks as graph kernel \cite{DBLP:conf/iclr/KipfW17}.
In this work, we consider the signal propagation will be different in different topological distances and to reduce the learning complexity, according to \cite{defferrard2016convolutional}, we use the following polynomial graph kernel $g_{v',v} (L )=\sum ^{K}_{k=0} \theta_{v',v,k} (I-L )^{k}$.

It follows that the intensity function is formalized as:
\begin{equation}
\label{eq:intensity_all}
\small
\begin{aligned}
 & \lambda _{v} (n,t)\\
= & \mu _{v} +\sum _{v'\in \mathbf{V}}\sum _{n'\in \mathbf{N}}\sum ^{K}_{k=0} 
\alpha _{v',v,k}\hat{A}_{n',n}^k\int_{t'\in \mathbf{T}_{t^{-}}} \kappa(t-t')dC_{v'} (n',t'),
\end{aligned}
\end{equation}
where $\hat{A}=D^{-1/2}AD^{-1/2}$ is the normalized adjacency matrix, and $\hat{A}^k_{n',n}$ is the $n',n$ entry of matrix $\hat{A}^k$. We further let $\alpha_{v',v,k}= \beta_{v',v}\theta_{v',v,k}$ represent the causal influence between $v'\to v$ from the $k$-hop paths and we can directly optimize the parameter $\alpha$ equivalently. The details of the derivation of the intensity function will be given in Appendix A. 
Intuitively, such an extended convolution operation measures the cumulative effect from all the past events that are causally related as well as topological close to node $n$ at time $t$.
In detail, $\hat{A}^k$ measures the accumulated effect from all length-$k$ closed walks in graph $\mathcal{G}_N$. 
$K$ refers to the farthest topological distance that the events can affect.
Therefore, the intensity of type $v'$ events at node $n'$ and time $t$ in Eq. \ref{eq:intensity_all} is the summation over all cause event types from history through all paths in different lengths.

In the following, we provide the theoretical analysis of the topological point process based on the work \cite{mogensen2020markov}. We first extend the notation of intensity to introduce the definition of local independence. For $S \subseteq \mathbf{V}$, suppose that $v \in S$, we have $\lambda_{v}^{S}(t)=E[\lambda_{v}(t)|\mathcal{F}^S(t)]$ according to the innovation theorem \cite{jacobsen2006point}. Then, the local independence of multivariate point processes is defined as follows: 
\begin{definition}[Local independence of multivariate point processes \cite{mogensen2020markov}]\label{def:local_ind}
Let $v,v',S \subseteq \mathbf{V}$. We say that the counting process $C_{v}$ is locally independent of $C_{v'}$ given $C_{S}$ if there exists an $\mathcal{F}^S(t)$-predictable
version of $\lambda_{v'}^{ S\cup \{v\}}(t)$. We use $v \not \to v' |S$ to denote that $C_v$ is locally independent of $C_v'$ given $C_S$, and else if $\lambda_{v'}^{S\cup \{v\}}(t)$ is not a version of $\lambda_{v'}^S(t)$, we write $v \to v' |S$.
\end{definition}
Intuitively that means that if a point process $C_v$ is locally independent of $C_{v'}$ given $C_S$ then the past up until time $t$ of $C_S$ has the same predictable information about $E(C_{v'}(t)|\mathcal{F}^{S \cup \{v\}}(t))$ as the past of $E(C_{v'}(t)|\mathcal{F}^{S}(t))$ until time $t$. Note that the local independence is generally not symmetric, and thus it naturally entails the causality among event types.
With Definition \ref{def:local_ind}, we are able to generalize the local independence to the topological point processes as follows:
\begin{definition}[Local independence of topological point processes]\label{def:local_ind_topo}
Let $v,v',S \subseteq \mathbf{V}$. We say that the counting process $C_{v}$ is locally independent of $C_{v'}$ given $C_{S}$ if there exists an $\mathcal{F}^{S,n}(t)$-predictable
version of $\lambda_{v'}^{\{S,n\}\cup \{v\}}(n',t)$ for all $n,n'\in \mathbf{N}$. We use $v \not \to v'|_{\mathcal{N}}S$ to denote that $C_v$ is locally independent of $C_v'$ given $C_S$ in the topological point process, and else if there exists $n,n' \in \mathcal{N}$ such that $\lambda_{v'}^{\{S,n\}\cup \{v\}}(n',t)$ is not a version of $\lambda_{v'}^{\{S,n\}}(n',t)$, we write $v \to v'|_{\mathcal{N}}S$.
\end{definition}
Definition \ref{def:local_ind_topo} illustrates the fact that if two event types is local independent in topological Hawkes processes, then it should also be independent across all nodes. Thus the following remark is straightforward.
\begin{remark}
\label{thm:thm2}
Given the causal structure $\mathcal{G}_V(\mathbf{V},\mathbf{E}_V)$ and $\mathcal{G}_N(\mathbf{N},\mathbf{E}_n)$, let $C$ be the topological Hawkes processes with the intensity function defined in Eq. \ref{eq:discrete_intensity}. Then $v'\rightarrow v \notin \mathbf{E}_V$ if and only if $\alpha _{v',v,k}\hat{A}_{n',n}^{k}=0$ for all $k\in \{0,...,K\}$, $n',n\in \mathbf{N}$ and $t\in[0,\infty)$.
\end{remark}

\noindent The complete proof of Remark \ref{thm:thm2} is also given in Appendix B. Remark \ref{thm:thm2} inspires a way to identify the causal relationship by detecting whether the impact $\alpha_{v',v,k}$ is zero or not.

\subsection{Likelihood of the Topological Hawkes Process}
Based on the above theoretical analysis, we know that it is crucial to constrain the sparsity to obtain a reasonable causal structure. Hence, we devise the objective function with sparsity constraint using BIC score under the likelihood framework.

Due to the event sequences being often collected within a period of time in many real-world applications, we mainly focus on the discrete time scenario in this work. In the discrete time scenario, the minimal time interval is $\Delta t$, $\mathbf{T}$ is reduced to the discrete set $\mathbf{T} =\{0,\Delta t,2\Delta t,...,T\}$, $dC_{v}(n,t)$ is reduced to the count of occurrence events of type $v$ event at node $n$ in time interval $[t-\Delta t, t]$. It follows that we can let $\mathbf{X} = \{X_{n,v,t}|n\in \mathbf{N},v \in  \mathbf{V},t \in \mathbf{T}\}$ denote a set of observations of $dC_{v}(n,t)$, and the integral operation in the continuous time domain in Eq.\ref{eq:intensity_all} is transformed into the sum operation in the discrete time domain. Given the set of the observations, we can estimate the intensity function as follows:  
\begin{equation}
\label{eq:discrete_intensity}
    \small
      \lambda _{v} (n,t)
    = \mu _{v} +\sum _{v'\in \mathbf{PA}_v}\sum _{n'\in \mathbf{N}}\sum ^{K}_{k=0} \alpha _{v',v,k} \hat{A}^{k}_{n',n}  \sum _{t'\in \mathbf{T}_{t^{-}}} \kappa ( t-t') X_{n',v',t'},
\end{equation}
where $\mathbf{PA}_v$ denotes the set of caused event types of $v$, i.e. the parents of $v$ in the causal graph $\mathcal{G}_V$.

Given Eq. \ref{eq:discrete_intensity}, the likelihood can be expressed as the function of $\mathcal{G}_V,\Theta$ and the topological structure $\mathcal{G}_N$ where $\Theta=(\mu_v,\alpha_{v',v,k})_{v',v\in \mathbf{V},k=0,\dots,K}$ contains all the parameters of the model. The log-likelihood is given in Eq. \ref{L} and the detail of the derivation will be provided in Appendix C.
\begin{equation}
\small
\label{L}
\begin{aligned}
 & L(\mathcal{G}_{V} ,\Theta ;\mathbf{X} ,\mathcal{G}_{N} )\\
= & \sum\limits _{v\in \mathbf{V}}\sum\limits _{t\in \mathbf{T}}\sum _{n\in \mathbf{N}} P\left( X_{n,v,t} |H^{\mathbf{PA}_{v}}_{t}\right)\\
= & \sum\limits _{v\in \mathbf{V}}\sum\limits _{t\in \mathbf{T}}\sum _{n\in \mathbf{N}} [-\lambda _{v} (n,t)\Delta t+X_{n,v,t}\log (\lambda _{v} (n,t))]+Const
\end{aligned}
\end{equation}
where $H_{t}^{\mathbf{PA}_v}$ denotes the set of historical events of the cause of the event type $v$, which occurred before time $t$. Intuitively, the likelihood shows the goodness of fit of the estimated parameters. Thus, it is plausible to obtain the confidence level of the estimated parameters using the likelihood ratio test \cite{vuong1989likelihood}. Note that the log-likelihood function will tend to produce excessive redundant causal edges. Thus, to enforce the sparsity, we introduce the Bayesian Information Criterion (BIC) penalty $\frac{p\log(m)}{2}$ into $L(\mathcal{G}_V, \Theta; \mathbf{X}, \mathcal{G}_N)$, where $p=|\mathbf{V}|+K|\mathbf{E}_V|$ is the number of parameters, and $m$ is the total number of events that have occurred in $\mathbf{T}$, according to \cite{mailan2017ssrn}. The new objective with BIC penalty is given as follows:
\begin{equation}
\label{L_B}
\small
  L_{B} (\mathcal{G}_V, \Theta ;\mathbf{X}, \mathcal{G}_N)
= L(\mathcal{G}_{V} ,\Theta ;\mathbf{X} ,\mathcal{G}_{N} ) -\frac{p\log( m)}{2},
\end{equation}
The BIC penalty is equivalent to the $\ell_0$-norm of parameters, where the number of parameters $p$ is the $\ell_0$-norm $|\Theta|_0$, and $\frac{\log(m)}{2}$ is the weight of $\ell_0$-norm.

\subsection{Sparse Optimization of the Likelihood}

In this section, we show how to optimize the above objective function with the $\ell_0$ sparsity constraints. Though various methods have been proposed for the sparse problem, e.g., the ADM4 method \cite{zhou2013learningsocial} uses a low-rank constraint on the causal graph, the MLE-SGL method \cite{xu2016learning} further introduces the sparse-group-lasso regularization, their performance still highly depends on the selection of the strength of sparsity regularization or the thresholds for pruning the structure which is not previously known.

To address this problem, we propose the sparse optimization scheme in two steps, such that the estimation and optimization steps are iteratively conducted to force to learn a sparse causal structure, i.e., \textit{estimation step}: $\sup_{\Theta}L_B(\mathcal{G}_V, \Theta;\mathbf{X},\mathcal{G}_N)$ and \textit{searching step}: $\max_{\mathcal{G}_V}\sup_{\Theta}L_{B}( \mathcal{G}_V ,\Theta ;\mathbf{X},\mathcal{G}_N)$, respectively. 
Briefly, in the estimation step, we estimate the BIC score with a given causal structure $\mathcal{G}_V$, and following \cite{lewis2011nonparametric,zhou2013learning}, we use an EM-based algorithm to estimate the parameters by optimizing $\sup_{\Theta}L_{B}( \mathcal{G}_V ,\Theta ;\mathbf{X},\mathcal{G}_N)$.
In the searching step, following \cite{tsamardinos2006max}, we use a Hill-Climbing algorithm to search for the best causal graph $\mathcal{G}_V$ with the highest BIC score. Different from the gradient-based method, the Hill-Climbing algorithm is a greedy algorithm and it searches the causal structure starting from an empty graph.
According to the work \cite{mailan2017ssrn}, given the large enough sample size, the consistency of BIC score holds asymptotically, i.e., BIC can choose a model with the minimum number of parameters that minimizes the Kullback-Leibler distance with true model. Then based on work \cite{chickering2002optimal}, such sparse optimization is able to find the true causal graph due to the consistency property of BIC score.

In detail, in the estimation step, we perform the EM-based algorithm to update the parameters iteratively. Note that if there is no causal relationship $v'\to v$ in $\mathcal{G}_V$, then the corresponding parameters $\alpha_{v',v,k}$ will be enforced to zero and will not be updated by EM. The details of the derivation of EM algorithm is given in Appendix D. Specifically, in the $i$-th step, the parameters are updated as follows:
\begin{equation}
\small
\label{update mu} 
\begin{aligned}
\mu ^{(i)}_{v} & =\frac{\sum\limits _{n\in \mathbf{N}}\sum\limits _{t\in \mathbf{T}} q^{\mu }_{n,v,t} X_{n,v,t}}{|\mathbf{N} ||\mathbf{T} |\Delta t} ,
\end{aligned}
\end{equation}

\begin{equation}
\small
\label{update alpha} 
\begin{aligned}
  \alpha ^{(i)}_{v',v,k}= \frac{\sum\limits _{n\in \mathbf{N}}\sum\limits _{t\in \mathbf{T}}\left[\sum _{n'\in \mathbf{N}}\sum _{t'\in \mathbf{T}_{t^{-}}} q^{\alpha }_{n,v,t} (n',v',t',k)\right] X_{n,v,t}}{\sum _{n\in \mathbf{N}}\sum\limits _{t\in \mathbf{T}}\left[\sum _{n'\in \mathbf{N}}\sum _{t'\in \mathbf{T}_{t^{-}}} \hat{A}^{k}_{n',n} \kappa ( t-t') X_{n',v',t'}\right] \Delta t} ,
\end{aligned}
\end{equation}

with 
\begin{equation*}
  \small
  q^{\mu }_{n,v,t} =\frac{\mu ^{(i-1)}_{v}}{\lambda ^{(i-1)}_{n,v} (t)},
\end{equation*}

\begin{equation*}
\small
q^{\alpha }_{n,v,t} (n',v',t',k)=\frac{\alpha ^{(i-1)}_{v',v,k} \hat{A}^{k}_{n',n} \kappa ( t-t') X_{n',v',t'}}{\lambda ^{(i-1)}_{v} (n,t)},
\end{equation*}
where
\begin{equation}
\small
\begin{aligned}
 & \lambda ^{(i-1)}_{v} (n,t)\\
= & \mu ^{(i-1)}_{v} +\sum _{v'\in \mathbf{PA}_{v}}\sum _{n'\in \mathbf{N}}\sum ^{K}_{k=0} \alpha ^{(i-1)}_{v',v,k} \hat{A}^{k}_{n',n}\sum _{t'\in \mathbf{T}_{t^{-}}} \kappa ( t-t') X_{n',v',t'}.
\end{aligned}
\end{equation}

\begin{algorithm}[tb]
	\caption{ Learning causal structure using THP with sparse optimization}
	\label{alg:search_alg}
	\textbf{Input}: Observation $\mathbf{X}$\\

    \textbf{Output}: $\mathcal{G}_V$, $\Theta$
	\begin{algorithmic}[1]

		\STATE $\mathcal{G}_V^*\leftarrow empty\ graph$, $L_B^*\leftarrow -\infty$, $L_B \leftarrow L_B^*-1$
		\STATE Initialize $\Theta^*$ randomly
		\WHILE{$L_B < L_B^*$}
		    \STATE $\langle \mathcal{G}_V,\Theta,L_B \rangle \leftarrow \langle \mathcal{G}_V^*,\Theta^*,L_B^* \rangle$
			\FOR{every $\mathcal{G}_V'\in \mathcal{V}(\mathcal{G}_V)$}
				\STATE Initialize $\Theta'$ randomly;
	
				\STATE \textbf{Repeat} Update $L_B'$ and $\Theta'$ via \ref{update mu} and \ref{update alpha} \textbf{Until} {convergence}
			\ENDFOR
			\STATE $\langle \mathcal{G}_V^*,\Theta^*,L_B^* \rangle \leftarrow \langle \mathcal{G}_V',\Theta',L_B' \rangle$ with largest $L_B'$
			
		\ENDWHILE
		\RETURN $\langle \mathcal{G}_V,\Theta \rangle$
	\end{algorithmic}
\end{algorithm}

To learn the best causal graph, a Hill-Climbing algorithm is proposed for searching the best $ \mathcal{G}_V $ with the highest $\sup_{\Theta}L_B(\mathcal{G}_V, \Theta;\mathbf{X},\mathcal{G}_N)$ that is optimized by the above EM algorithm (Line 7 of Algorithm \ref{alg:search_alg}). Note that EM might not converge to the global optimum, we can simply try different initial points of the parameters and choose the best one. In detail, as shown in Algorithm \ref{alg:search_alg}, let $\mathcal{V}(\mathcal{G}_V)$ denote the vicinity of the current causal graph structure, which is obtained by performing a one time operation on $\mathcal{G}_V$ with an edge added, deleted, or reversed operation. Then, at each iteration, we evaluate each the causal structure $\mathcal{G}'$ in the vicinity $\mathcal{V}(\mathcal{G}_V)$ to find optimum causal structure $\mathcal{G}_V^*$ with the highest score (Line 5-9 of Algorithm \ref{alg:search_alg}). Finally, to start over, set $\mathcal{G}_V=\mathcal{G}_V^*$ (Line 4 of Algorithm \ref{alg:search_alg}), repeat until the score no longer increase. In addition, $K$ is chosen based on a prior knowledge in algorithm \ref{alg:search_alg}, it is also legitimate to select $K$ with the highest BIC score.

The time complexity of Algorithm \ref{alg:search_alg} is listed below. 
The worst-case computational complexity of our proposed method is $\mathcal{O}\left( Km^{3} E\frac{|\mathbf{V}|( |\mathbf{V}|+1)}{2}\right)$ and the best-case computational complexity is $\mathcal{O}\left( Km^{3} |\mathbf{V}|^{2}\frac{\lceil E/|\mathbf{V}|\rceil (\lceil E/|\mathbf{V}|\rceil +1)}{2}\right)$.
The complexity analysis can be conducted on the estimation step and the maximization step respectively. In the estimation step, the computational complexity mainly depends on the EM algorithm. By using the constant iterative step of EM, given a causal graph $\mathcal{G}_V$, the complexity is $\mathcal{O}(Km^3e)$, where $K$ represents the farthest topological distance considered in the intensity function, $m$ is the number of occurring events and $e$ denotes the number of edges in $\mathcal{G}_V$. Note that if $\kappa$ satisfies $\kappa(t_1+t_2)=\kappa(t_1)\kappa(t_2)$, the complexity of EM decreases to $\mathcal{O}(Km^2e)$ because the intensity at each occurred event can be computed using only the intensity from the previous occurred event. In fact, in our implementation, the complexity is $\mathcal{O}(Km^2e)$ because of the exponential kernel. For general purposes, the following analysis does not include the special kernel. It is worth noting that the computational complexity of EM algorithm can be further improved by methods like stochastic approximation \cite{cappe2009line}.

In the searching step, because the initial graph is empty. That is the number of edges will start at $e=1$, then the time complexity highly depends on the number of parents at each event type. With the assumption of consistency score, the total steps can be considered as the number of edges that need to be added, which is denoted by $E$. Then, the worst time complexity is the case that all edges are concentrated in certain event types. Note that each step in the searching will need to apply the estimation step, thus we have $\mathcal{O}\left(\frac{E}{|\mathbf{V}|}\sum ^{|\mathbf{V}|}_{e=1} |\mathbf{V}|Km^{3} e\right) =\mathcal{O}\left( Km^{3} E\frac{|\mathbf{V}|( |\mathbf{V}|+1)}{2}\right)$. Note that the complexity of adding one edge in the worst case is $\mathcal{O}(Km^3|\mathbf{V}|e)$ because we only need to update the score on the type that the parents changed and all the scores on other types will be cached. Similarly, for the best case, each causal relationship is evenly distributed to each event type such that the number of parent of each type is $\lceil \frac{E}{|\mathbf{V}|} \rceil$, and we have the best case time complexity $\mathcal{O}\left( |\mathbf{V}|\sum ^{\lceil E/|\mathbf{V}|\rceil }_{e=1} |\mathbf{V}|Km^{3} e\right) =\mathcal{O}\left( Km^{3} |\mathbf{V}|^{2}\frac{\lceil E/|\mathbf{V}|\rceil (\lceil E/|\mathbf{V}|\rceil +1)}{2}\right)$.

\begin{figure*}[t]
	\centering
	\subfigure[Sensitivity to Range of $\alpha$]{
		\includegraphics[width=0.45\textwidth]{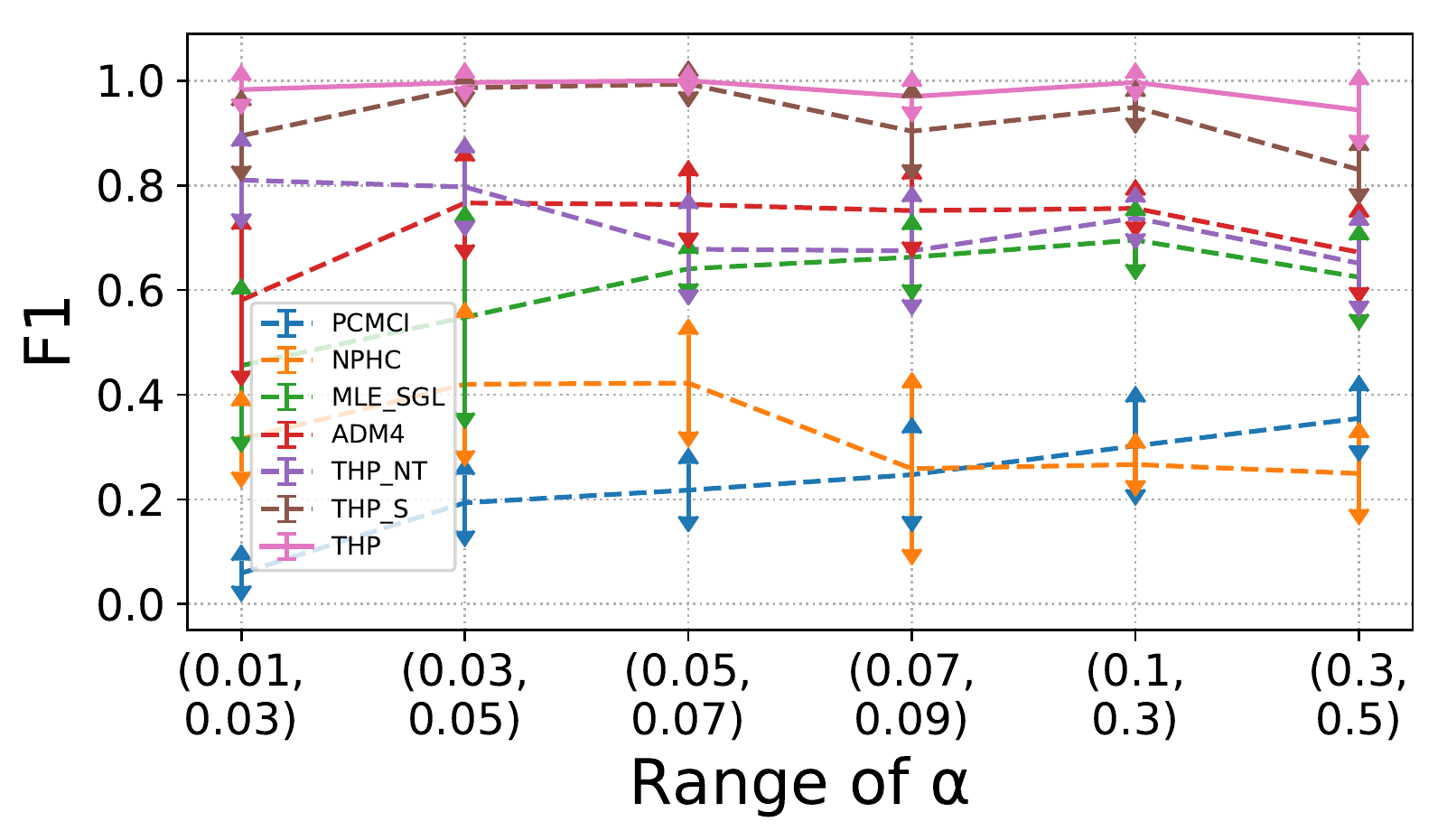}
		\label{f2:a}
	}
	\subfigure[Sensitivity to Range of $\mu$]{
	\includegraphics[width=0.45\textwidth]{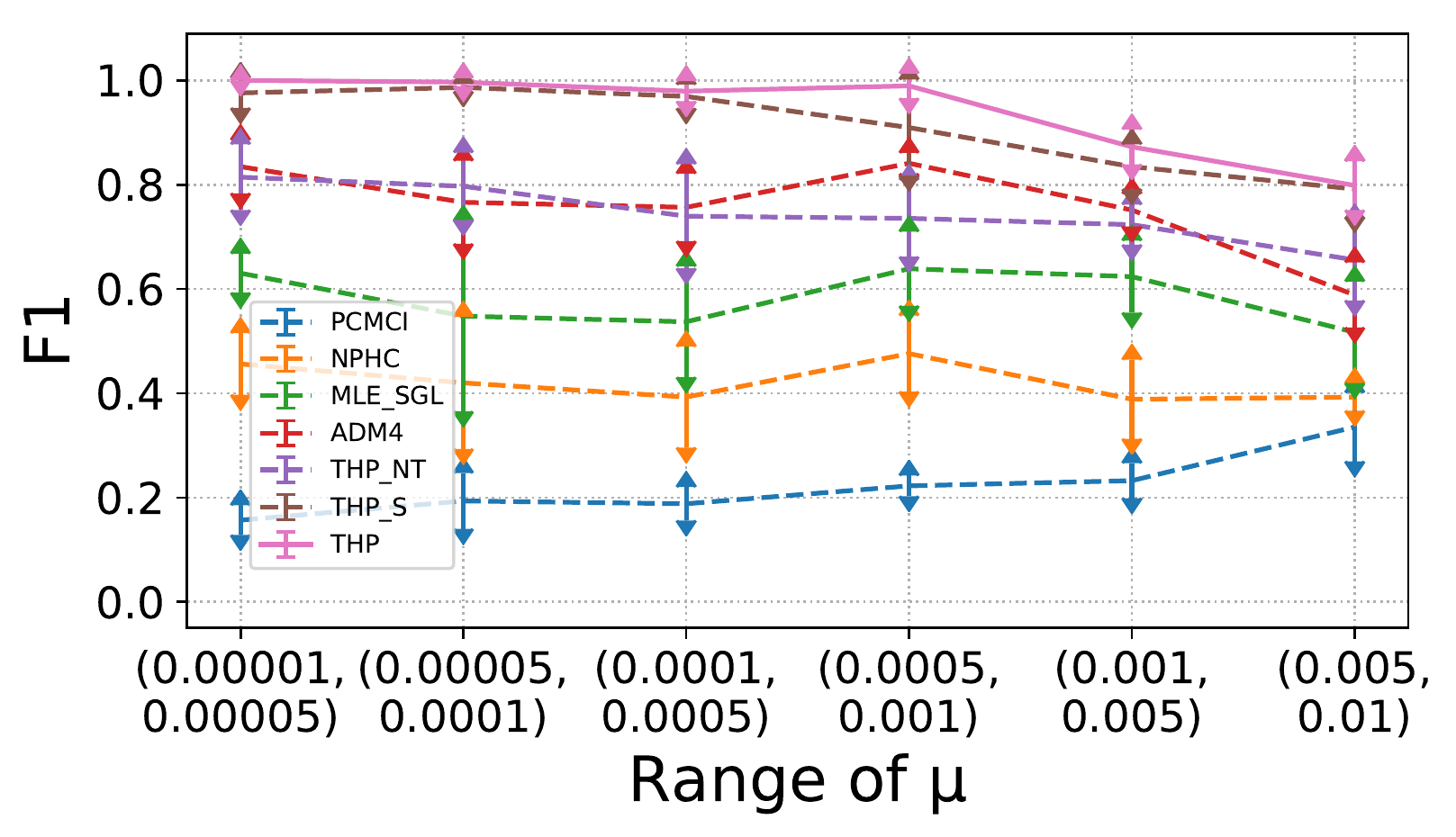}
	\label{f2:b}
}
	\subfigure[Sensitivity to Sample Size]{
	\includegraphics[width=0.45\textwidth]{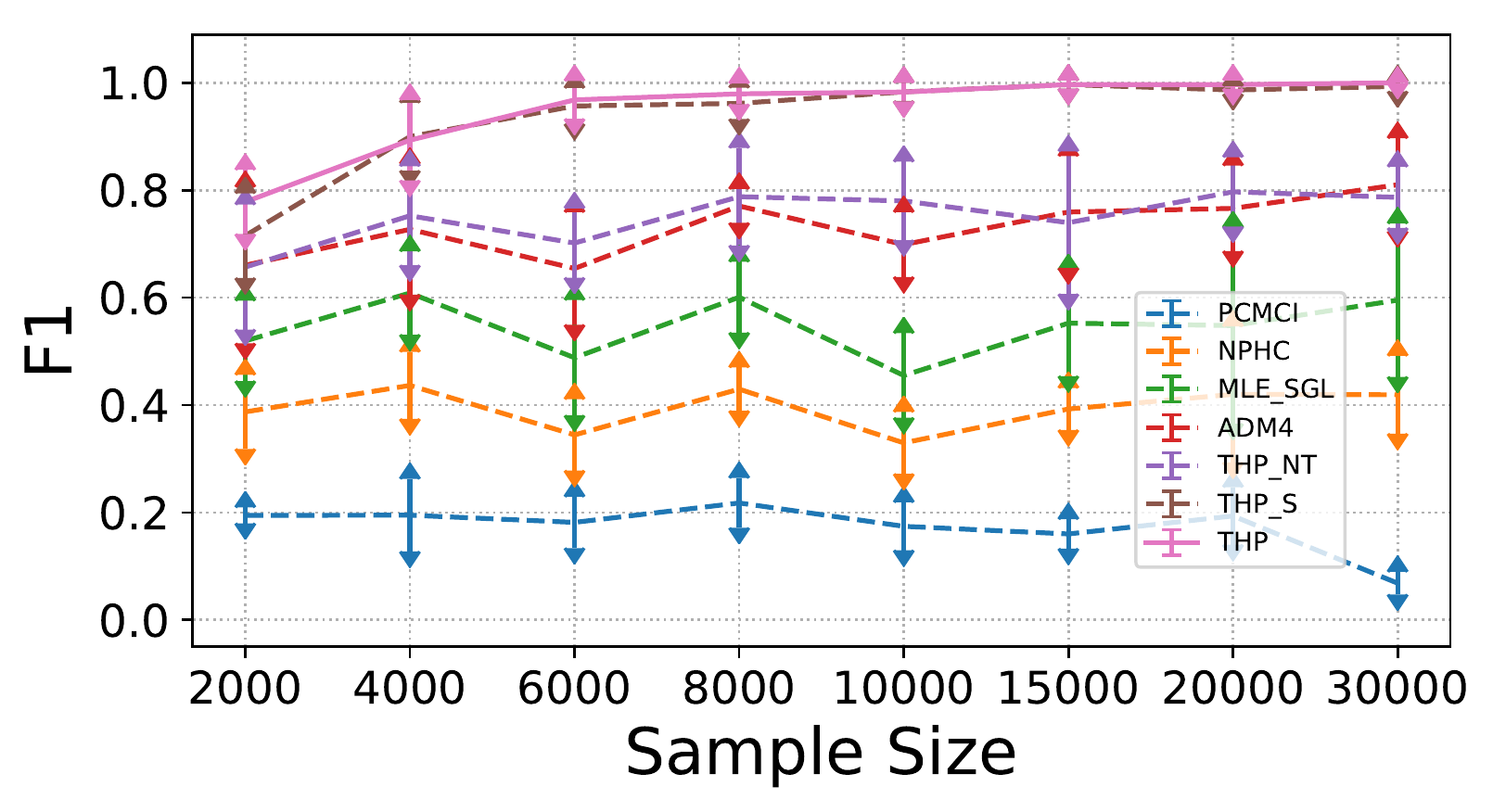}
	\label{f2:c}
}
	\subfigure[Sensitivity to Avg. Indegree]{
	\includegraphics[width=0.45\textwidth]{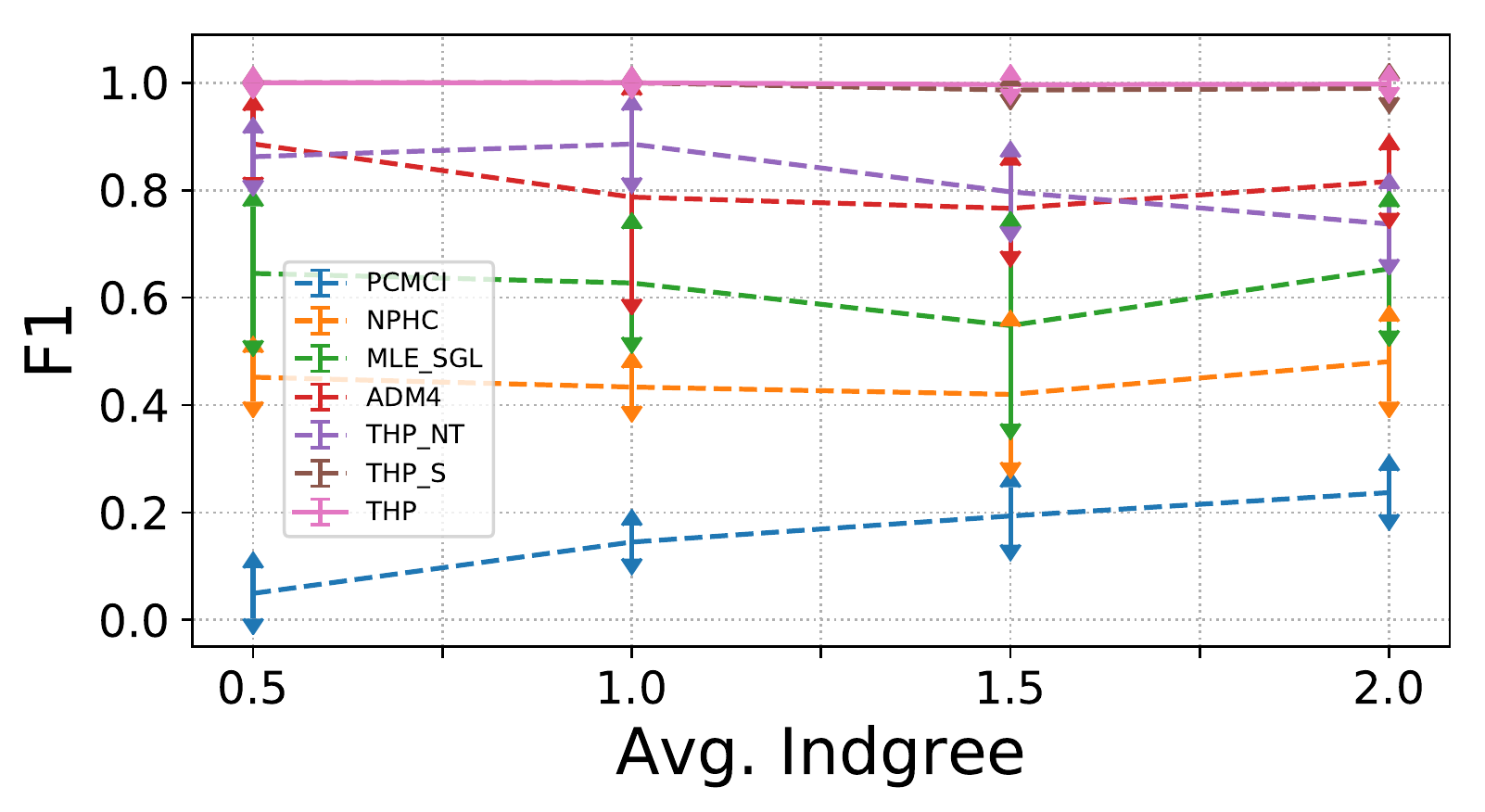}
	\label{f2:d}
}
	\subfigure[Sensitivity to Num. of Event Types]{
	\includegraphics[width=0.45\textwidth]{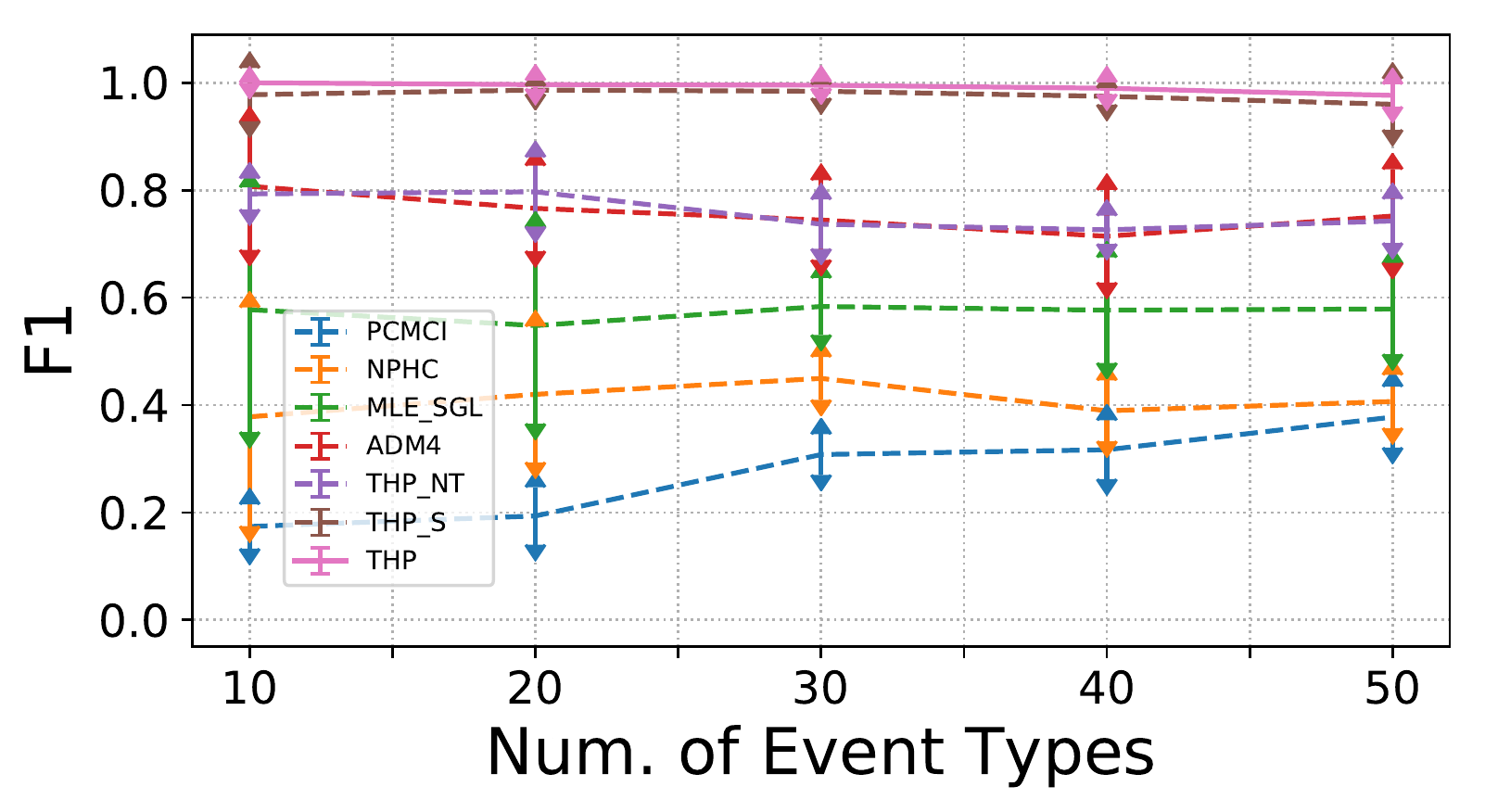}
	\label{f2:e}
}
	\subfigure[Sensitivity to Num. of Nodes]{
	\includegraphics[width=0.45\textwidth]{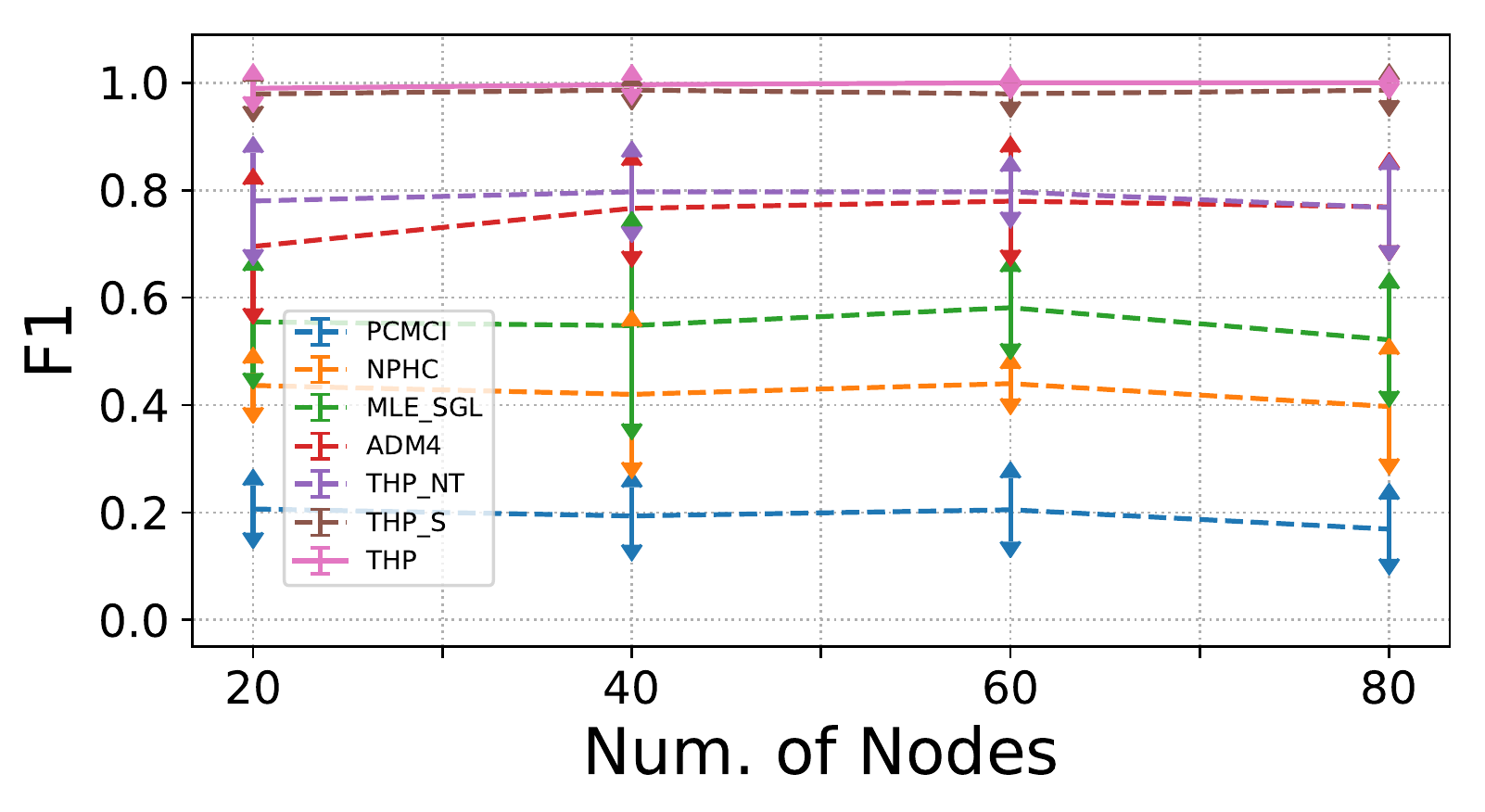}
	\label{f2:f}
}
	\caption{F1 in the Sensitivity Experiments with Variance}	
	\label{f2}
\end{figure*}

\begin{figure*}[t]
	\centering
	\subfigure[Sensitivity to Range of $\alpha$]{
		\includegraphics[width=0.45\textwidth]{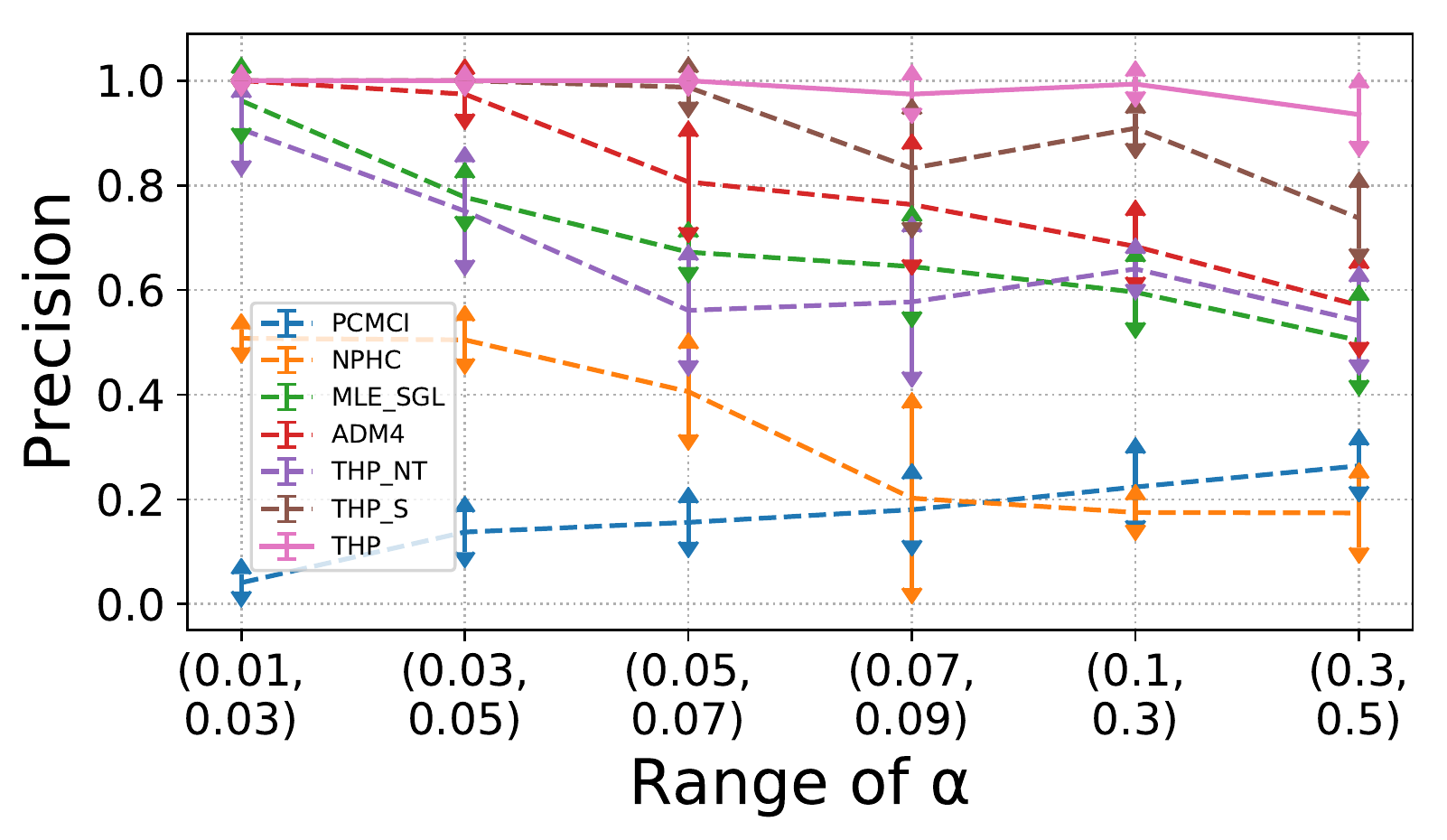}
		\label{f3:a}
	}
	\subfigure[Sensitivity to Range of $\mu$]{
	\includegraphics[width=0.45\textwidth]{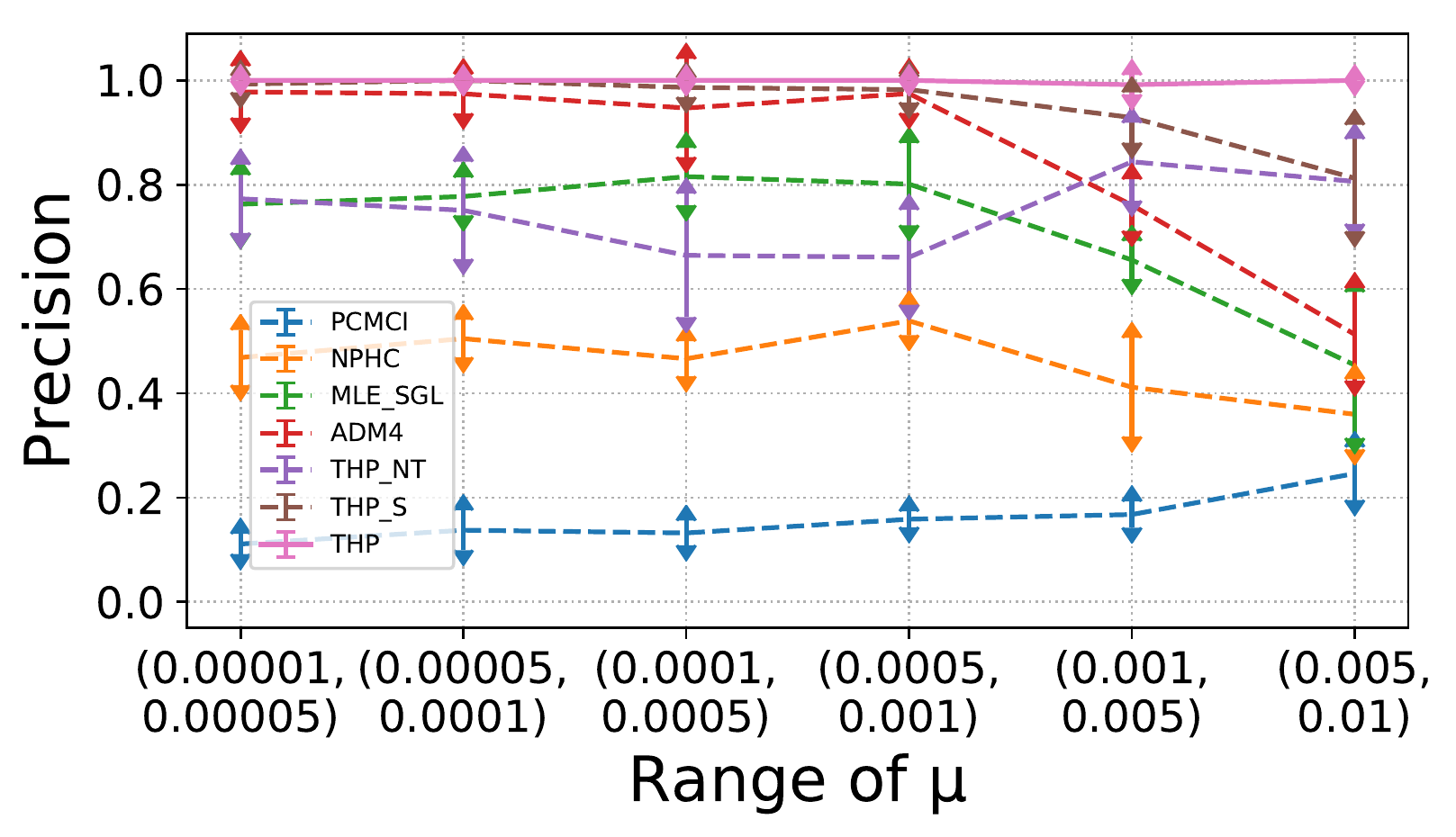}
	\label{f3:b}
}
	\subfigure[Sensitivity to Sample Size]{
	\includegraphics[width=0.45\textwidth]{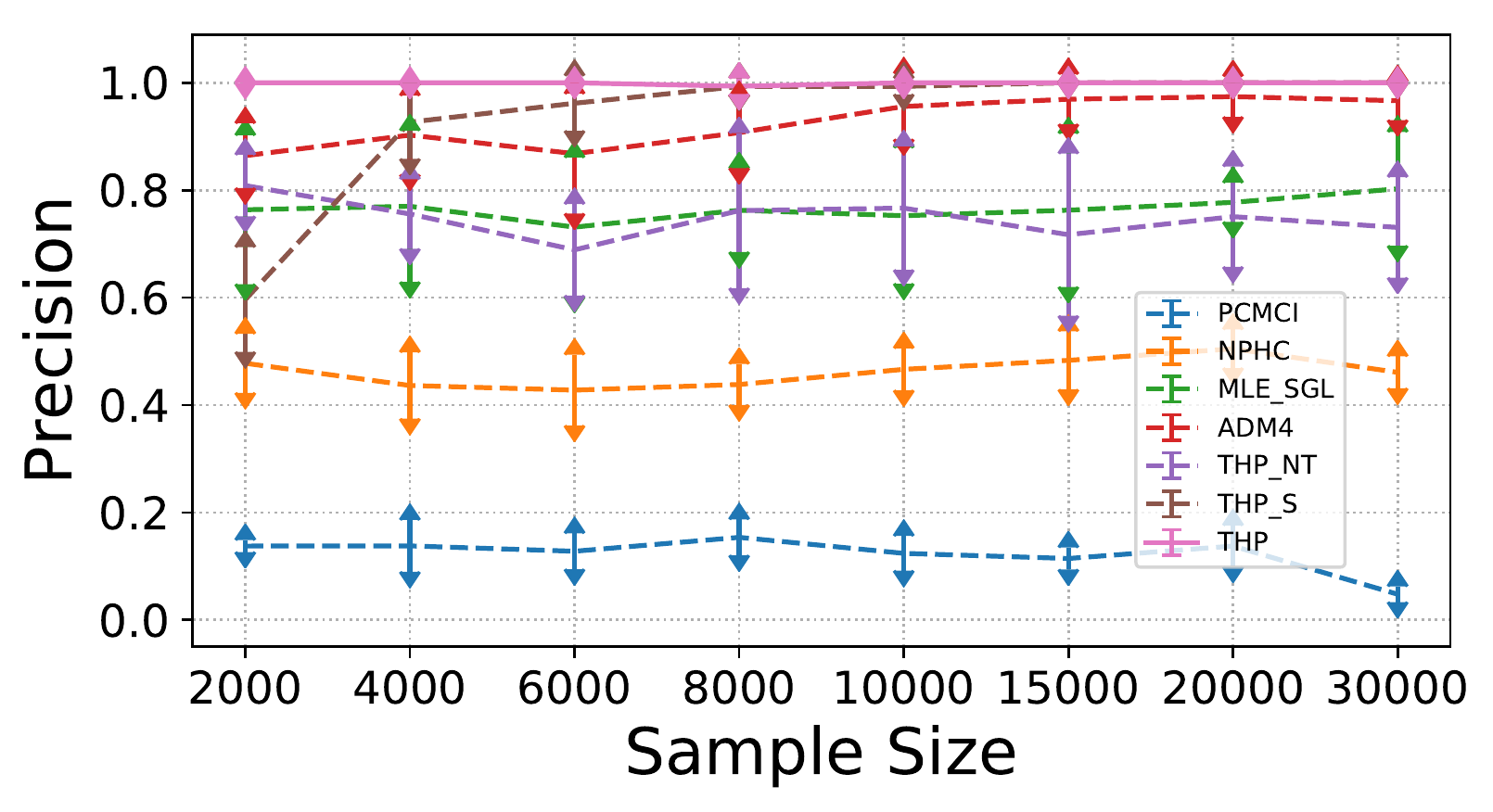}
	\label{f3:c}
}
	\subfigure[Sensitivity to Avg. Indegree]{
	\includegraphics[width=0.45\textwidth]{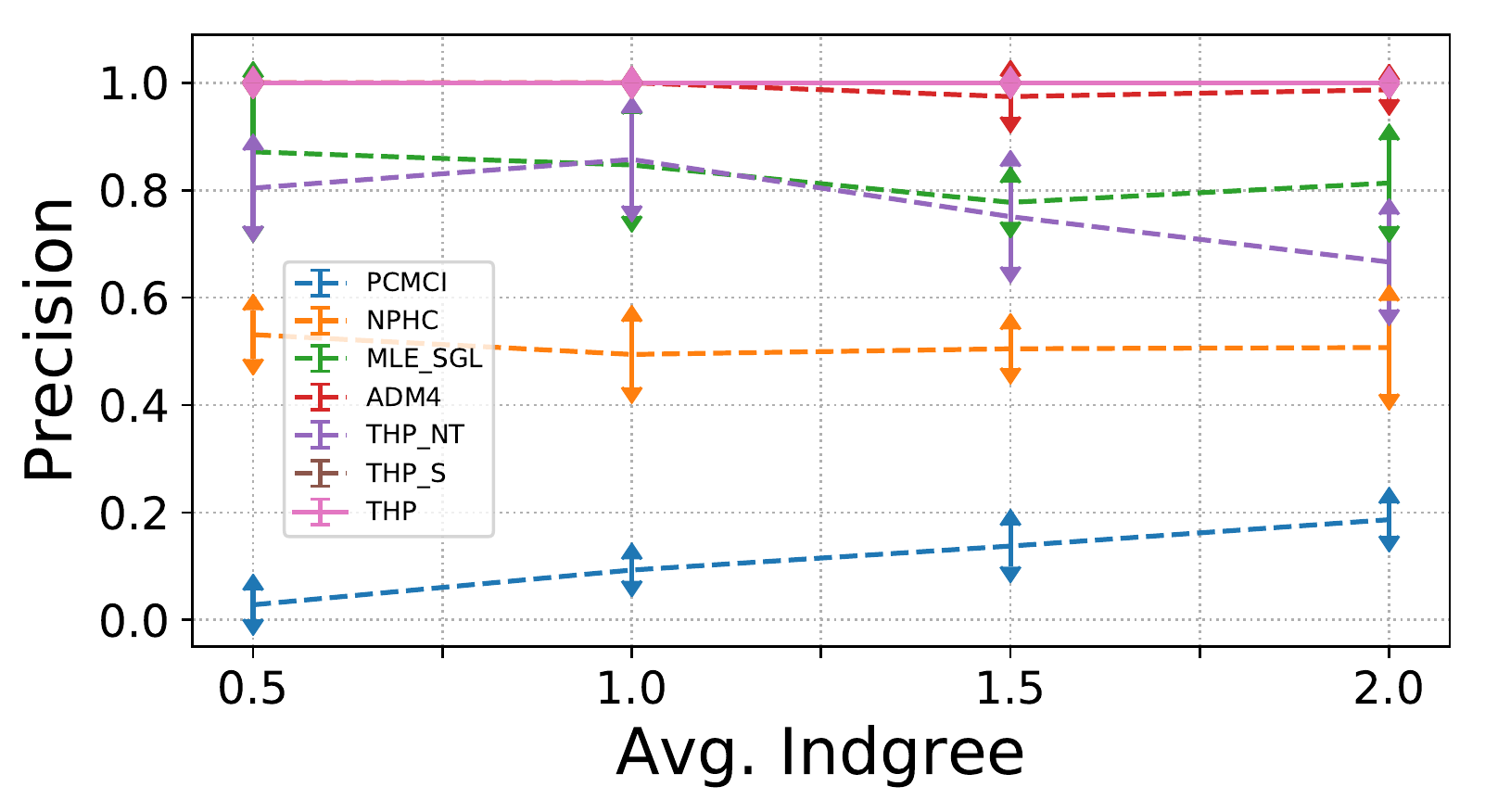}
	\label{f3:d}
}
	\subfigure[Sensitivity to Num. of Event Types]{
	\includegraphics[width=0.45\textwidth]{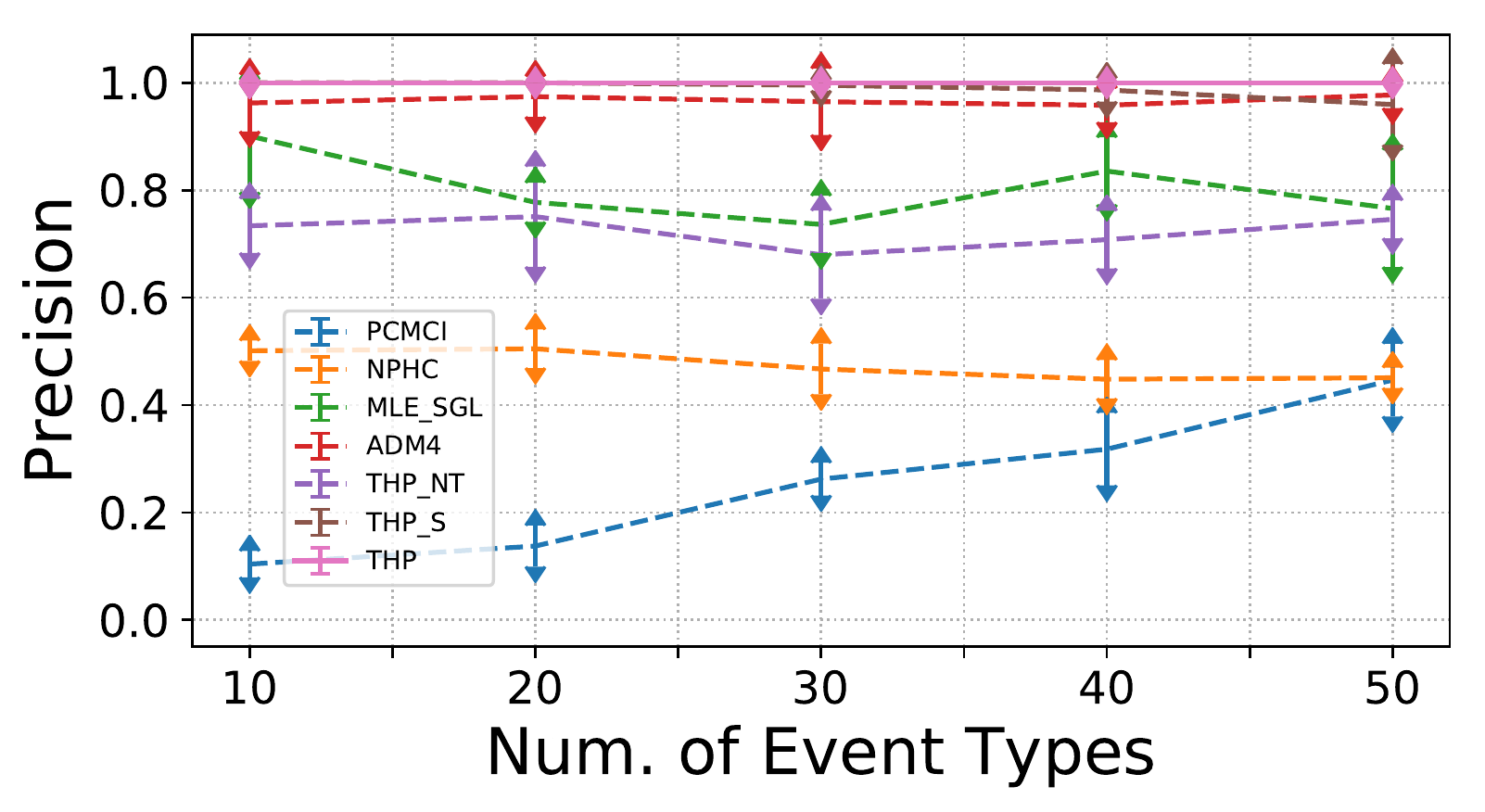}
	\label{f3:e}
}
	\subfigure[Sensitivity to Num. of Nodes]{
	\includegraphics[width=0.45\textwidth]{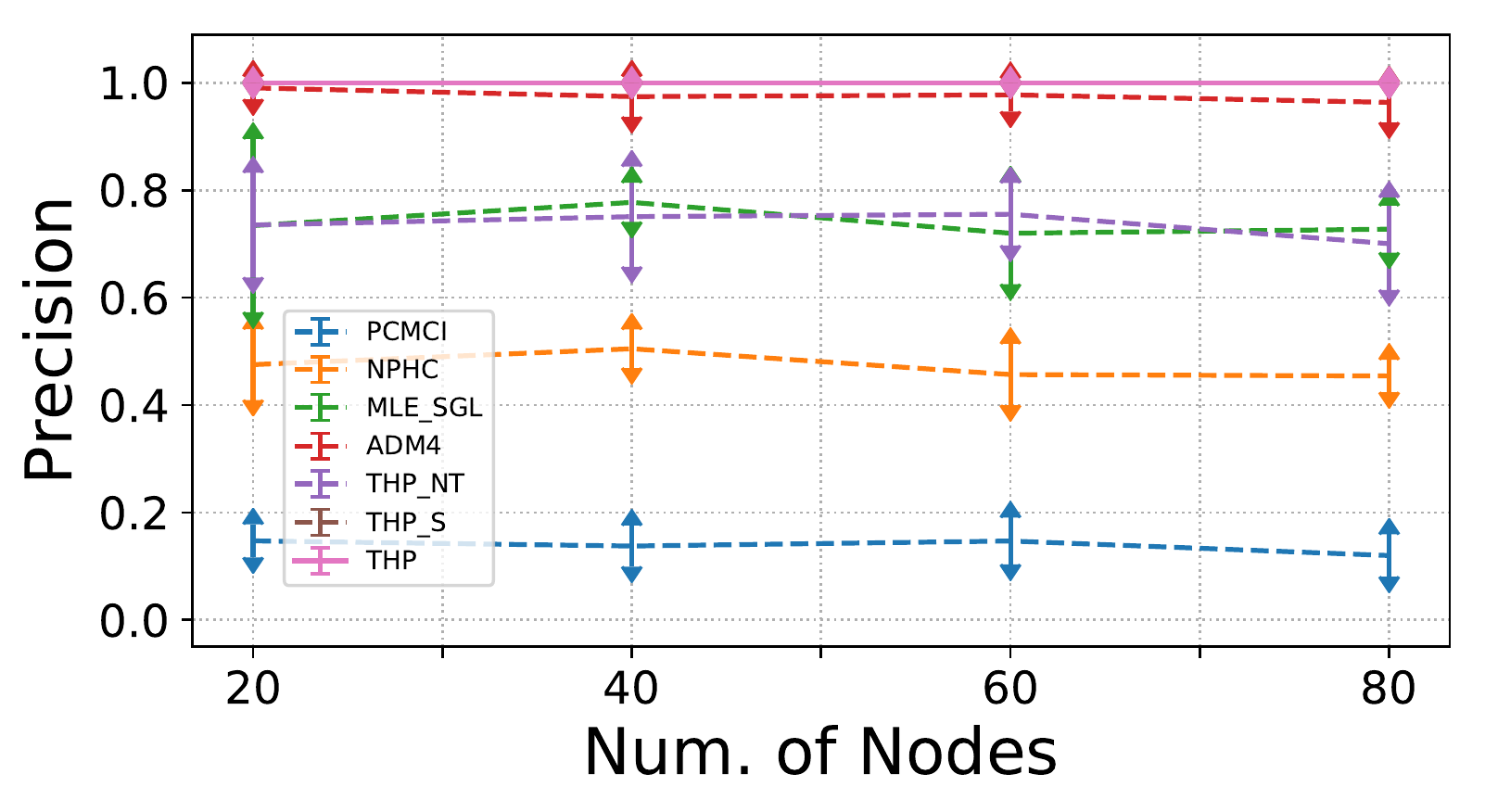}
	\label{f3:f}
}
	\caption{Precision in the Sensitivity Experiments with Variance}	
	\label{f3}
\end{figure*}

\begin{figure*}[t]
	\centering
	\subfigure[Sensitivity to Range of $\alpha$]{
		\includegraphics[width=0.45\textwidth]{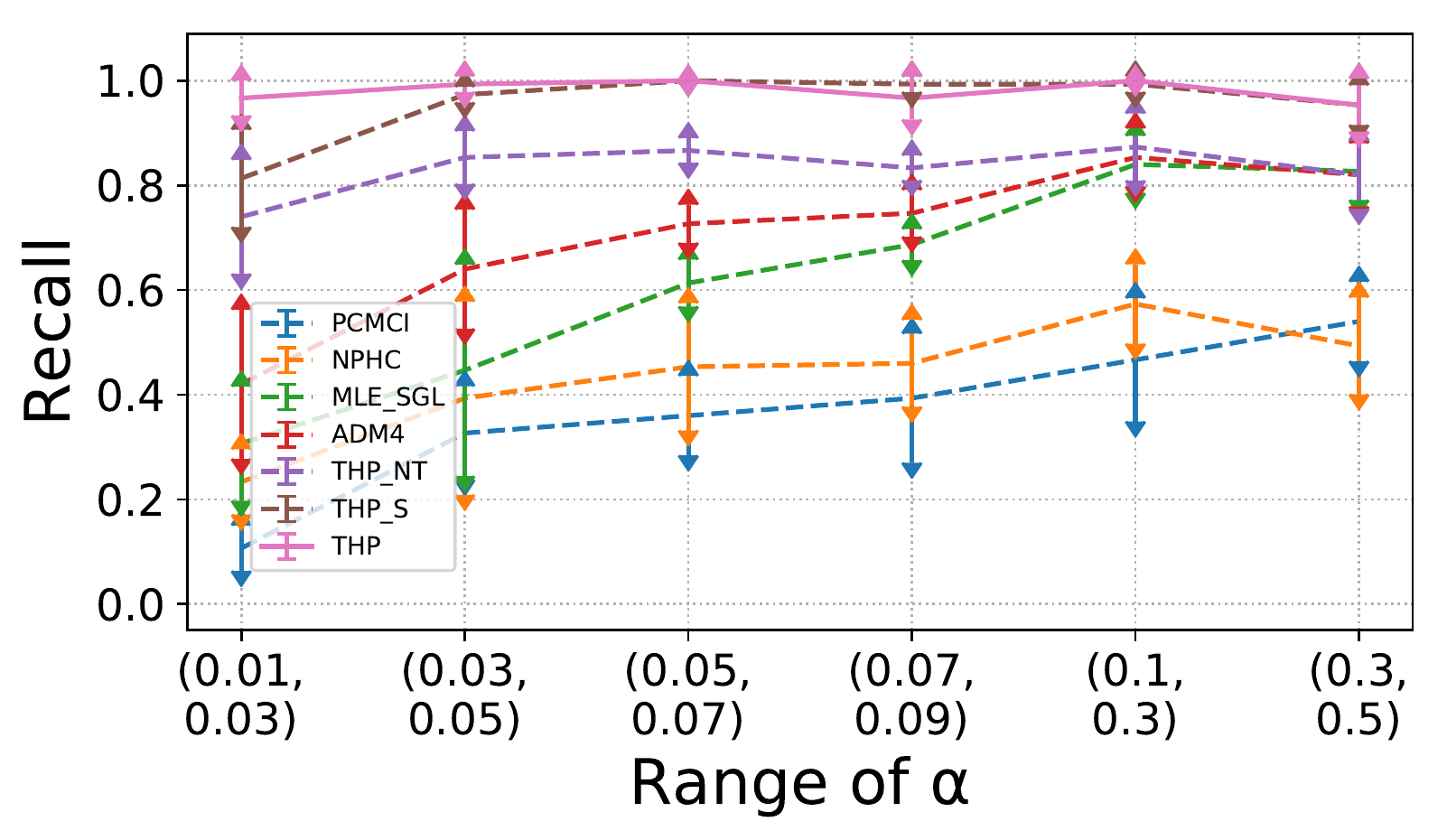}
		\label{f4:a}
	}
	\subfigure[Sensitivity to Range of $\mu$]{
	\includegraphics[width=0.45\textwidth]{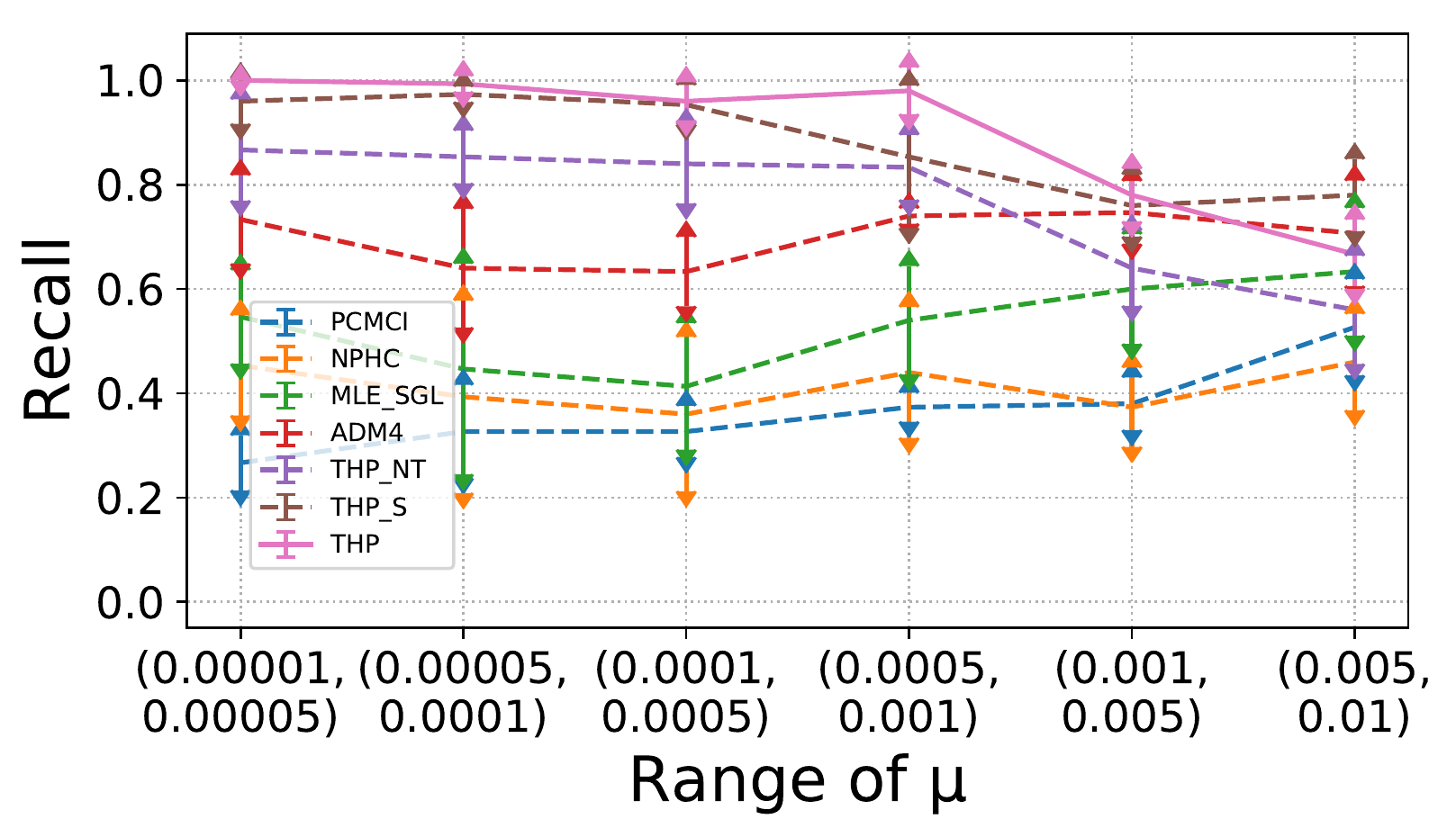}
	\label{f4:b}
}
	\subfigure[Sensitivity to Sample Size]{
	\includegraphics[width=0.45\textwidth]{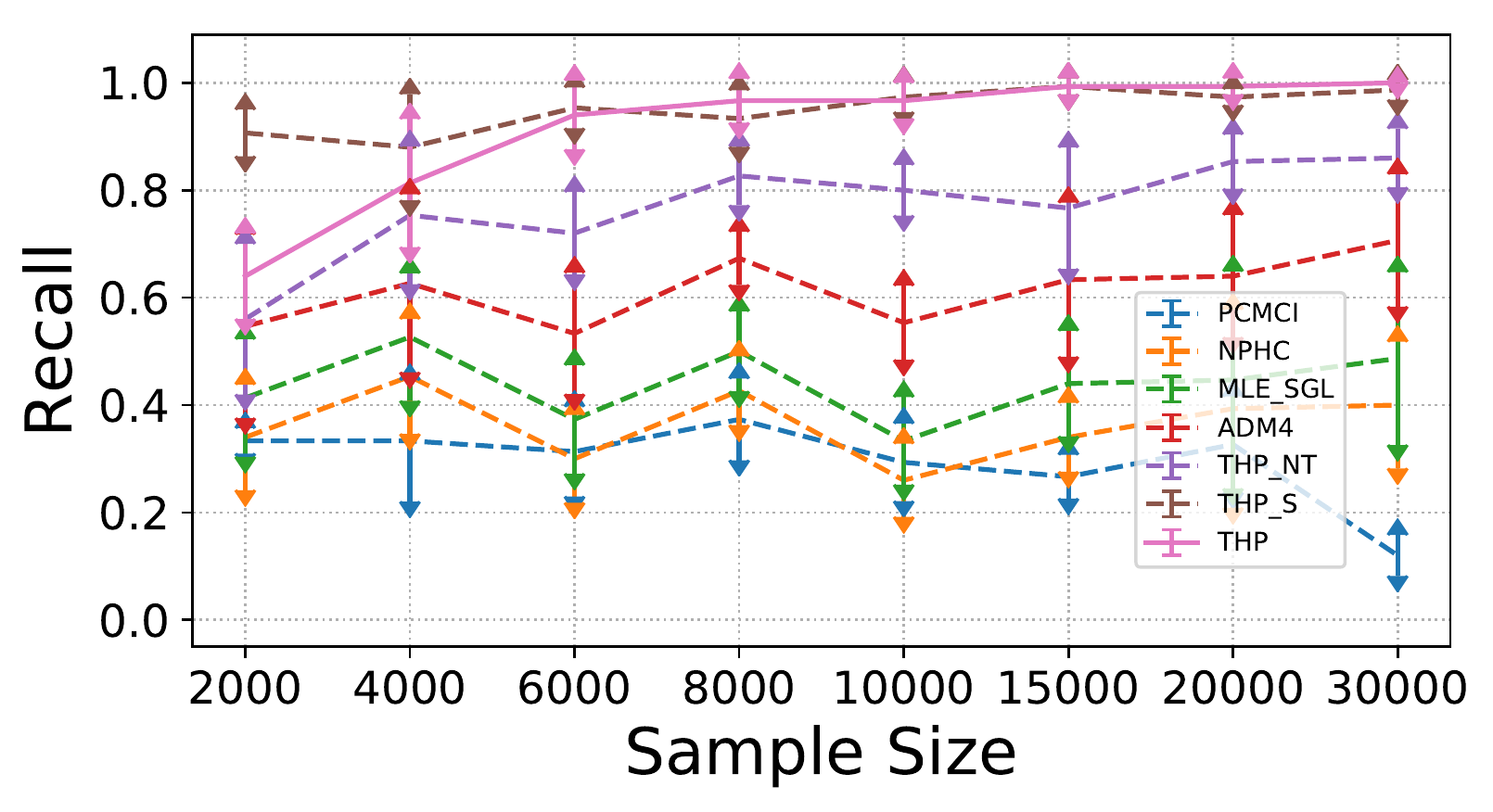}
	\label{f4:c}
}
	\subfigure[Sensitivity to Avg. Indegree]{
	\includegraphics[width=0.45\textwidth]{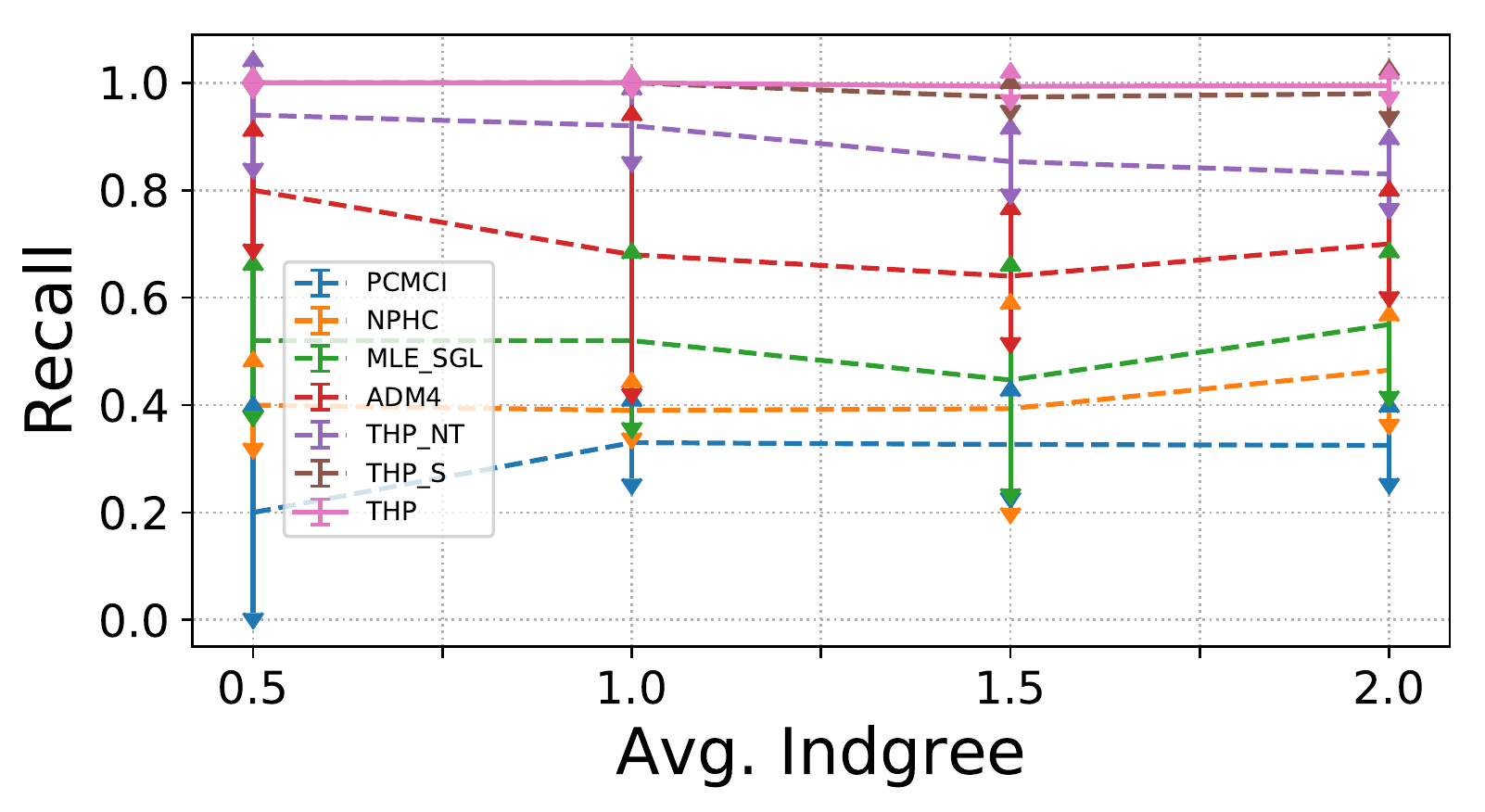}
	\label{f4:d}
}
	\subfigure[Sensitivity to Num. of Event Types]{
	\includegraphics[width=0.45\textwidth]{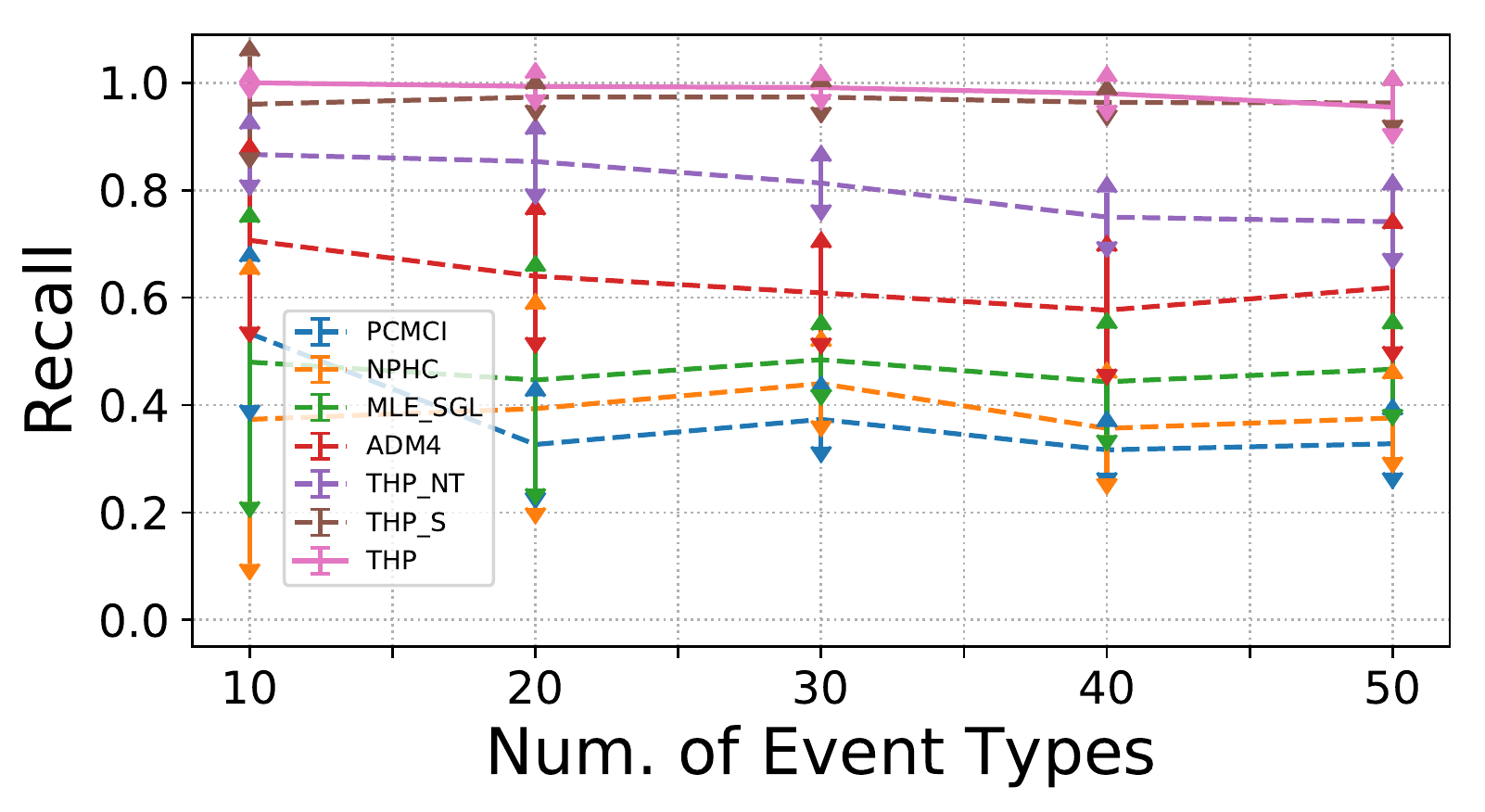}
	\label{f4:e}
}
	\subfigure[Sensitivity to Num. of Nodes]{
	\includegraphics[width=0.45\textwidth]{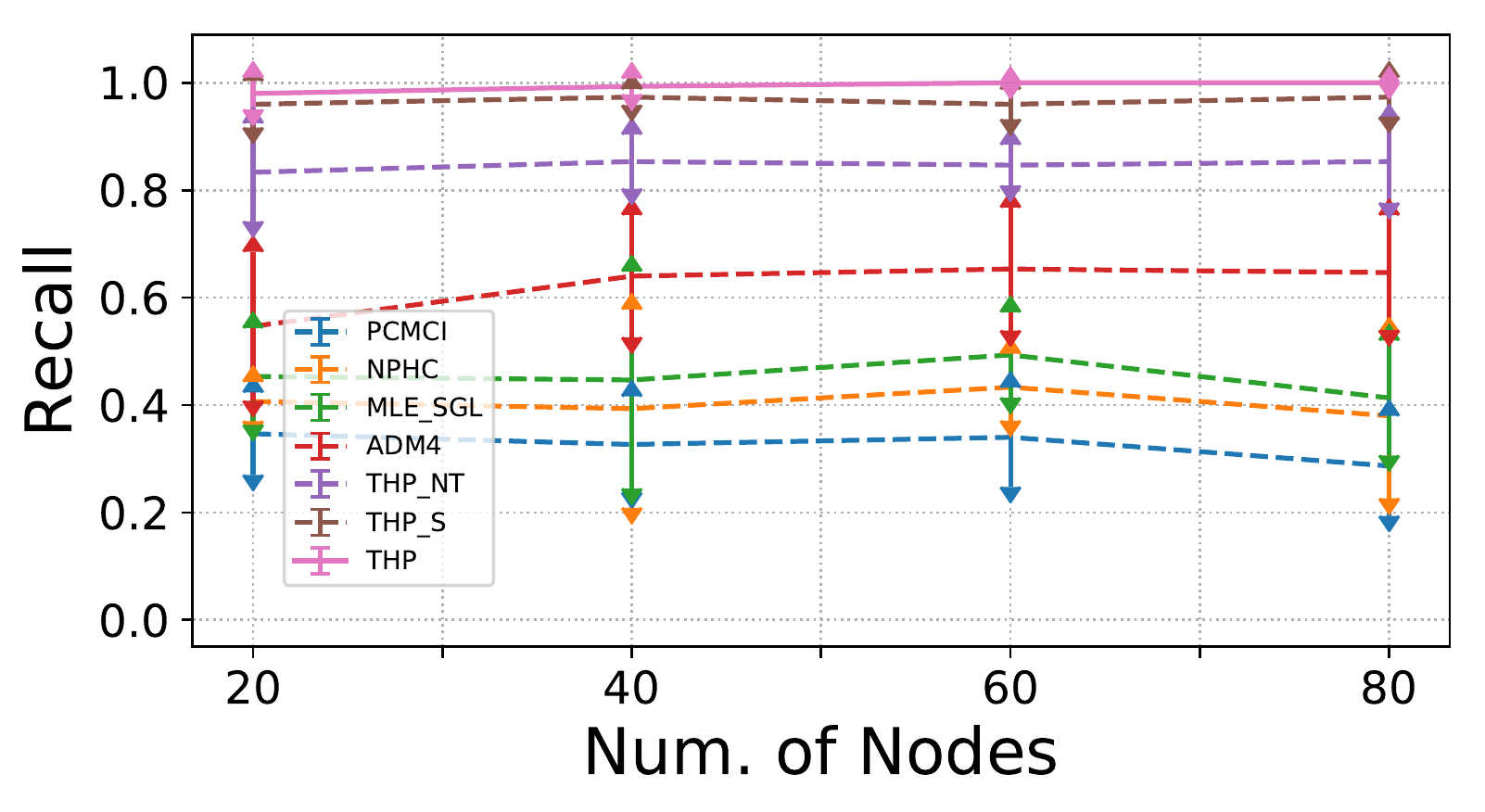}
	\label{f4:f}
}
	\caption{Recall in the Sensitivity Experiments with Variance}	
	\label{f4}
\end{figure*}

\begin{figure}[t]
    \centering
    \includegraphics[width=0.45\textwidth]{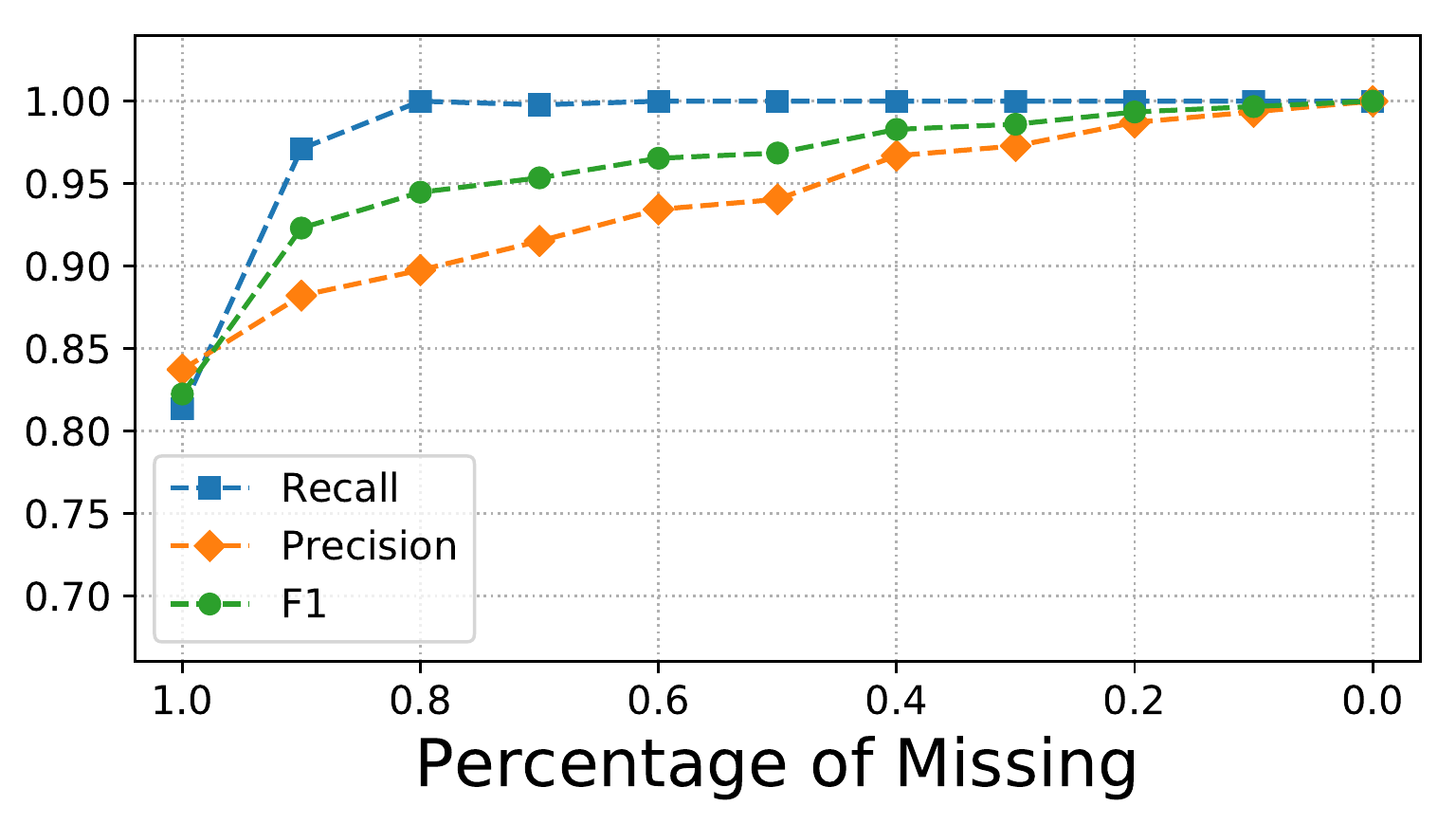}
    \caption{Performance of THP on Data with Partially Observed Topology}
    \label{fig:del_topo_exp}
\end{figure}

\begin{figure*}[tb]
	\centering
	\subfigure[$k=0$]{
		\includegraphics[height=0.3\textwidth]{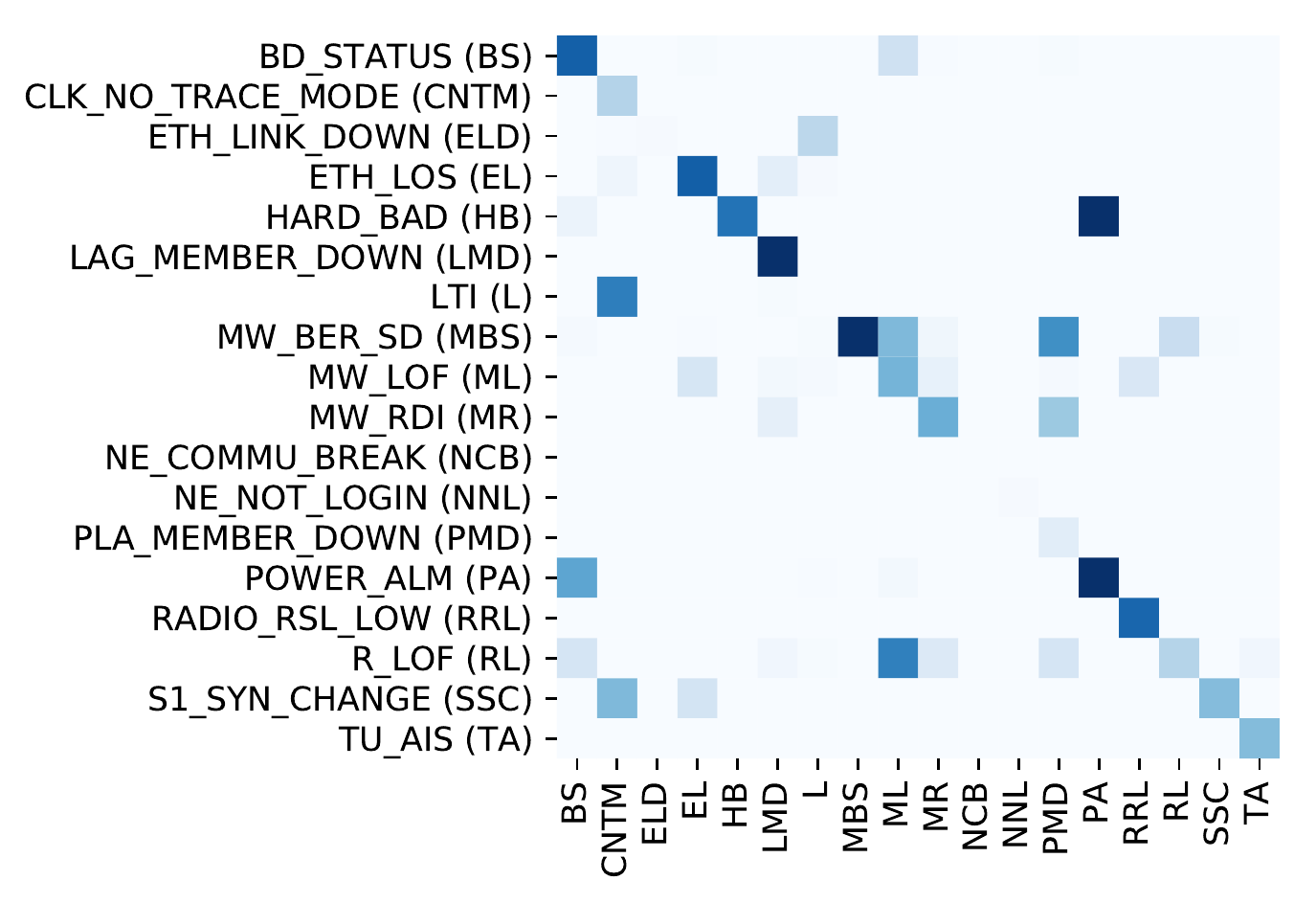}
		\label{f5:a}
	}
	\subfigure[$k=1$]{
	\includegraphics[height=0.3\textwidth]{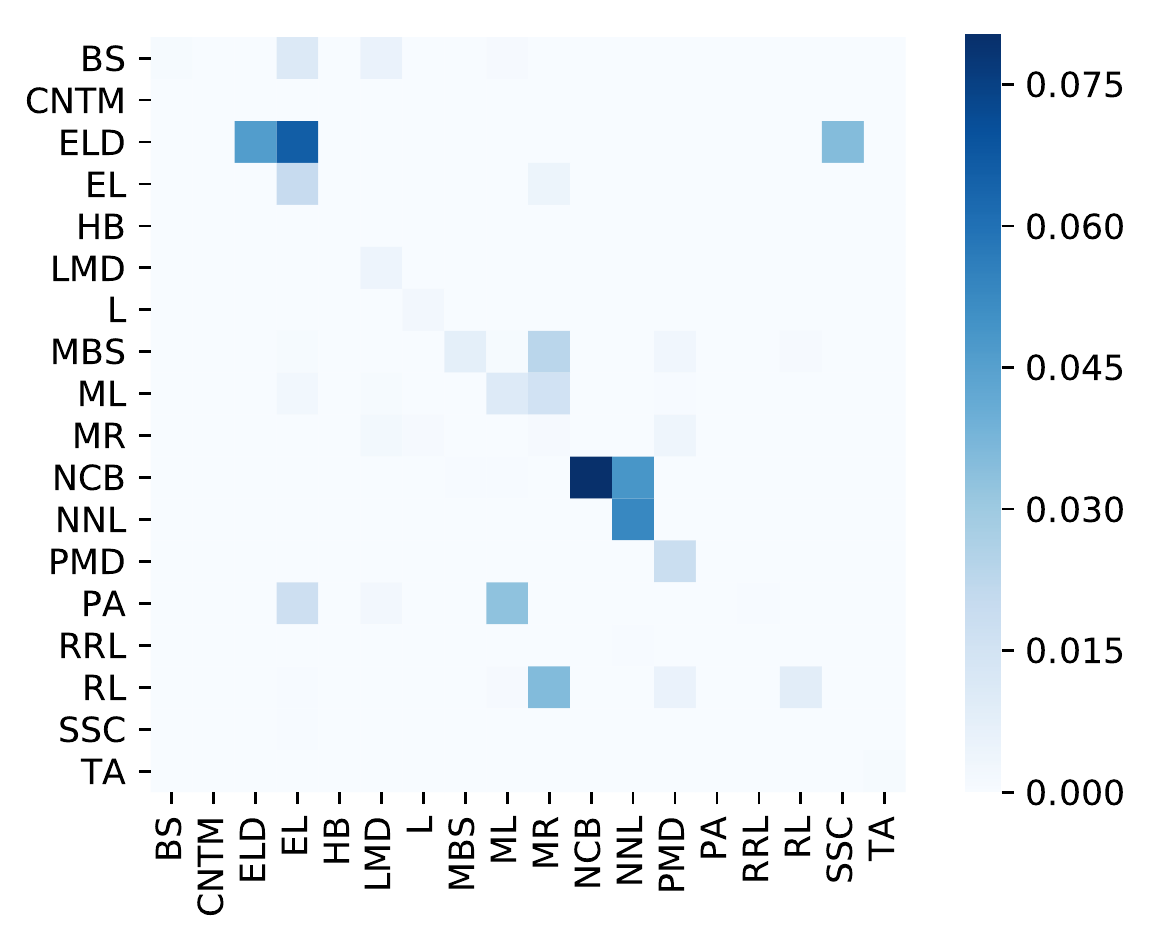}
	\label{f5:b}
}
	\subfigure[$k=2$]{
	\includegraphics[height=0.3\textwidth]{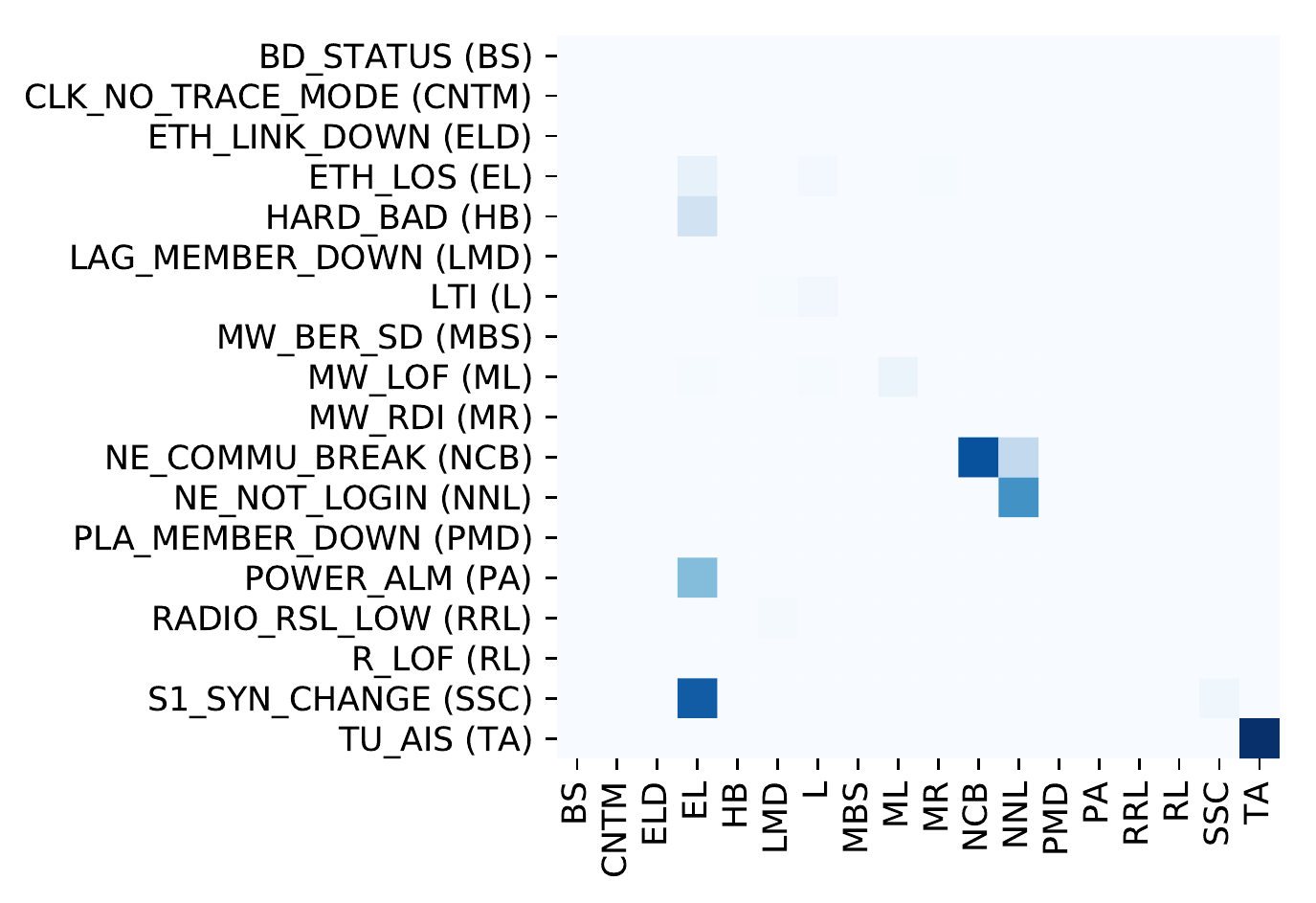}
	\label{f5:d}
}
	\subfigure[$k=3$]{
	\includegraphics[height=0.3\textwidth]{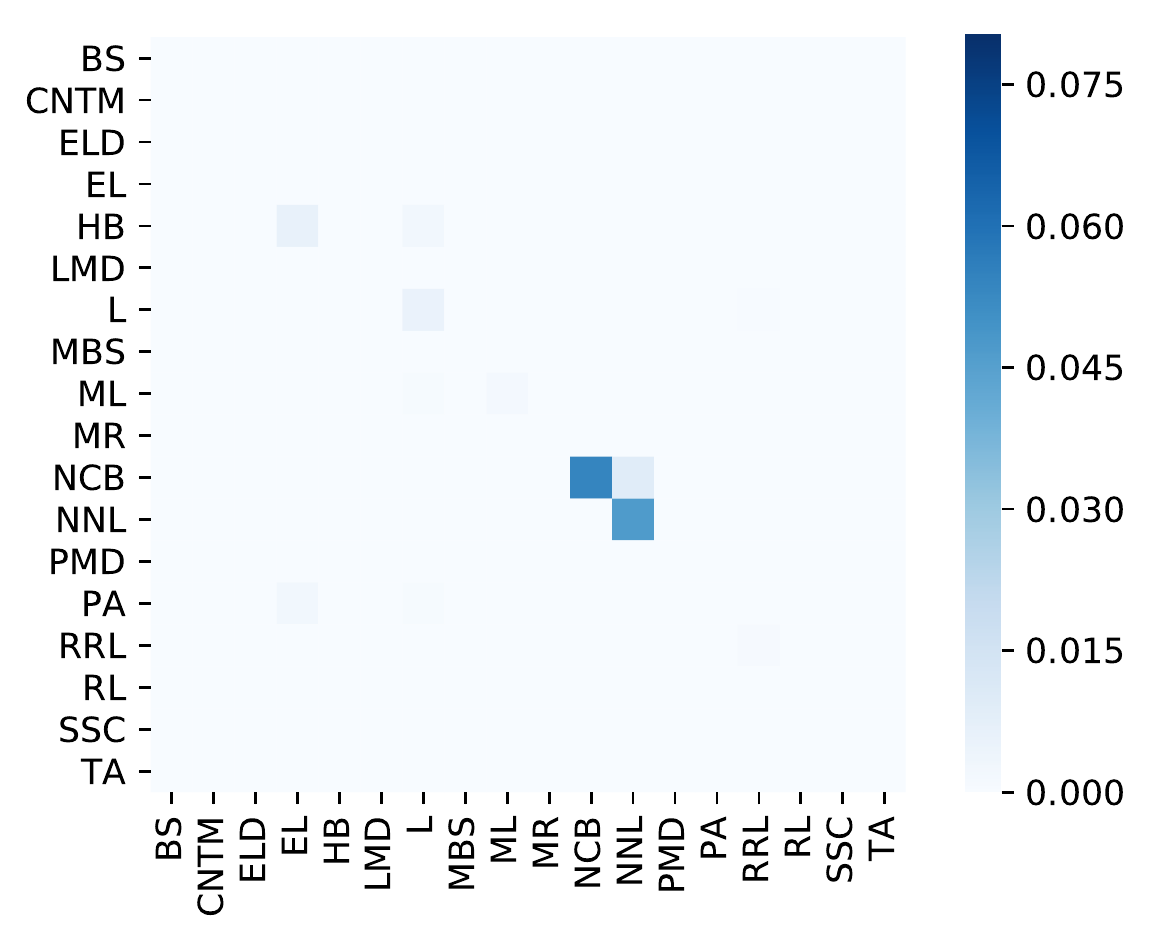}
	\label{f5:e}
}
	\caption{Causal strength $\alpha_{v',v,k}$ in different topological distance.}	
	\label{f5}
\end{figure*}

\section{Experiments}
In this section, we test the proposed THP and the baselines on both the synthetic data and the real-world metropolitan cellular network alarm data. The baselines include PCMCI \cite{runge2019detecting}, NPHC \cite{achab2017uncovering}, MLE-SGL \cite{xu2016learning}, and ADM4 \cite{zhou2013learningsocial}. To study the performance of each component, we also develop two variants of THP, namely THP\_NT and THP\_S. THP\_NT is a variant of THP by removing the topology information. THP\_S is another variant of THP by removing the sparse optimization schema but using the causal strength threshold to determine the causal structure following the work of ADM4 and MLE-SGL. The details of these baselines are provided in Appendix F. In all the following experiments, Recall, Precision, and F1 are used as the evaluation metrics:

\begin{equation}
\begin{aligned}
\text{Precision} & =\frac{\text{Number of correctly predicted\ edges}}{\text{Number of edges of predicted structure}} ,\\
\text{Recall} & =\frac{\text{Number of correctly predicted\ edges}}{\text{Number of edges of real structure}} ,\\
\text{F1} & =\frac{2\times \text{Precision} \times \text{Recall}}{\text{Precision} +\text{Recall}} .
\end{aligned}
\end{equation}

\begin{table}[t]
\caption{F1 Score of Gaussian Form Impact Functions Exp.}
	\centering
\begin{tabular}{l|rrrrr}
\toprule
\multirow{2}{*}{} & \multicolumn{5}{c}{\textbf{Different ranges of mean $a$ of Gaussian kernel}} \\ \cmidrule{2-6} 
                  & \multicolumn{1}{c}{${10\pm 5}$} & \multicolumn{1}{c}{${50\pm 5}$} & \multicolumn{1}{c}{${100\pm 5}$} & \multicolumn{1}{c}{${300\pm 5}$} & \multicolumn{1}{c}{${500\pm 5}$} \\ \midrule
\textbf{PCMCI}                & 0.16564                         & 0.13264                         & 0.09949                          & 0.11459                          & 0.07710                          \\
\textbf{NPHC}                 & 0.50427                         & 0.14149                         & 0.17947                          & 0.17519                          & 0.17443                          \\
\textbf{MLE\_SGL}             & 0.84155                         & 0.77917                         & 0.74377                          & 0.62706                          & 0.48050                          \\
\textbf{ADM4}                 & 0.84948                         & 0.76788                         & 0.75685                          & 0.63494                          & 0.50030                          \\
\textbf{THP\_NT}              & 0.82613                         & 0.78761                         & 0.73534                          & 0.74195                          & 0.69190                          \\
\textbf{THP}                  & \textbf{1.00000}                & \textbf{1.00000}                & \textbf{0.94635}                 & \textbf{0.86645}                 & \textbf{0.84192}                 \\ \bottomrule
\end{tabular}
\label{tab:gaussian}
\end{table}

\begin{table}[t]
\caption{F1 Score of Uniform Impact Functions Exp.}
\begin{tabular}{l|rrrrr}
\toprule
& \multicolumn{5}{c}{\textbf{Different ranges of starting point $b$ of Uniform kernel}} \\ \cmidrule{2-6} 
\textbf{}         & \multicolumn{1}{c}{${10\pm 5}$} & \multicolumn{1}{c}{${50\pm 5}$} & \multicolumn{1}{c}{${100\pm 5}$} & \multicolumn{1}{c}{${300\pm 5}$} & \multicolumn{1}{c}{${500\pm 5}$} \\ \midrule
\textbf{PCMCI}                & 0.10654                         & 0.10353                         & 0.15884                          & 0.12872                          & 0.11195                          \\
\textbf{NPHC}                 & 0.40558                         & 0.13917                         & 0.10605                          & 0.10886                          & 0.12691                          \\
\textbf{MLE\_SGL}             & 0.77463                         & 0.80044                         & 0.79564                          & 0.61279                          & 0.41919                          \\
\textbf{ADM4}                 & 0.84997                         & 0.81520                         & 0.79549                          & 0.60286                          & 0.43569                          \\
\textbf{THP\_NT}              & 0.80308                         & 0.83881                         & 0.82670                          & 0.68227                          & 0.61930                          \\
\textbf{THP}                  & \textbf{1.00000}                & \textbf{0.99661}                & \textbf{0.91402}                 & \textbf{0.88879}                 & \textbf{0.87519}                 \\ \bottomrule
\end{tabular}
\label{tab:uniform}
\end{table}

\subsection{Synthetic Data}

\paragraph{Data Generation}
We use the following four steps to generate the multi-type event sequences with a topological graph: 1) randomly generate a directed causal graph $\mathcal{G}_V$ and the undirected topological graph $\mathcal{G}_N$; 2) generate the events in the root of types (i.e., there are no parents for these root types) using the Poisson process; 3) generate the other type of events according to the causal structure and the topological graph with randomly generated parameters $\alpha, \mu$. $\alpha$ denotes the causal strength in the intensity function and $\mu$ is the base intensity. 
\paragraph{Experiments setting} We conduct the sensitivity study on the synthetic data and the ablation study of the proposed method. In the sensitivity study, 
the generated parameters are fixed with default settings and traversed one by one as shown in Fig. \ref{f2}. The default settings are listed below, the number of nodes = $40$, the average degree = $1.5$, the number of event types = $20$, the sample size (number of events) = $20000$, the range of $\mu_v=[0.00005,0.0001]$, and the range of $\alpha_{v',v}=[0.03,0.05]$. 
Note that the range of these parameters is selected based on real-world data.
The F1, Recall, Precision results of the sensitivity experiments above of are given in Fig. \ref{f2}, \ref{f3}, \ref{f4}, respectively.
Moreover, to test the sensitivity of the kernel type in the impact function, we generate the data using Gaussian kernel:
\begin{equation*}
 \phi_{v',v}( t) =\begin{cases}
\alpha_{v',v}\frac{1}{Z}\exp\left( - \Vert t-a\Vert ^{2}\right) & t> 0\\
0 & t\leqslant 0
\end{cases}
\end{equation*}
and uniform kernel: 
\begin{equation*}
 \phi_{v',v}( t) =\begin{cases}
\alpha_{v',v}/scale & b< t< b+scale\\
0 & \text{otherwise}
\end{cases}
\end{equation*}
where $Z$ is the normalizing constant of Gaussian distribution with standard deviation $4$, and the $scale$ of the Uniform kernel is set to $4$. By uniformly choosing the kernel location in different ranges, the experiment results are given in Table \ref{tab:gaussian} and Table \ref{tab:uniform}.

In the ablation study, we aim to test the effectiveness of the sparsity constrain and the topological component in THP. Specifically, the sparsity constraint is tested through THP\_S in which the sparsity constraint is removed. For the topological part, we test THP when there is missing or partially observed topological information. The completely missing case is given by THP\_NT while the partially missing case is given by randomly deleting the topological edges of $\mathcal{G}_N$. The results of the partially missing case are given in Figure \ref{fig:del_topo_exp}.

\paragraph{Comparison with the Baselines}
In all experiments, THP achieves the best results compared with other baselines. Among the baselines, ADM4 and MLE\_SGL are relatively better compared with NPHC and PCMCI. The reason is that ADM4 and MLE\_SGL have the sparse constraint but others do not, which shows the importance of the sparse constraint. 

\paragraph{Ablation Study}
Overall, THP is generally better than both THP\_S and THP\_NT, which the necessity of introducing the topological structure and the sparse optimization.
Though THN\_S does not use the sparse optimization, it is still better than other baselines, which verify the effectiveness of THP.

Furthermore, Fig . \ref{fig:del_topo_exp} shows how the performance of THP changes over the percentage of missing edges in the topological structure under the default setting. We can see that the topological structure plays an important role in precision but has a relatively small impact on recall as 80\% of missing is enough to produce $1$ recall. The reason is that missing topological edges would introduce unobserved confounders leading to redundant causal edges, which will decrease the precision. As the percentage of missing decreases, the precision increase which shows the necessity of considering topological structure.

\begin{figure}[tb]
    \centering
               \includegraphics[width=0.45\textwidth]{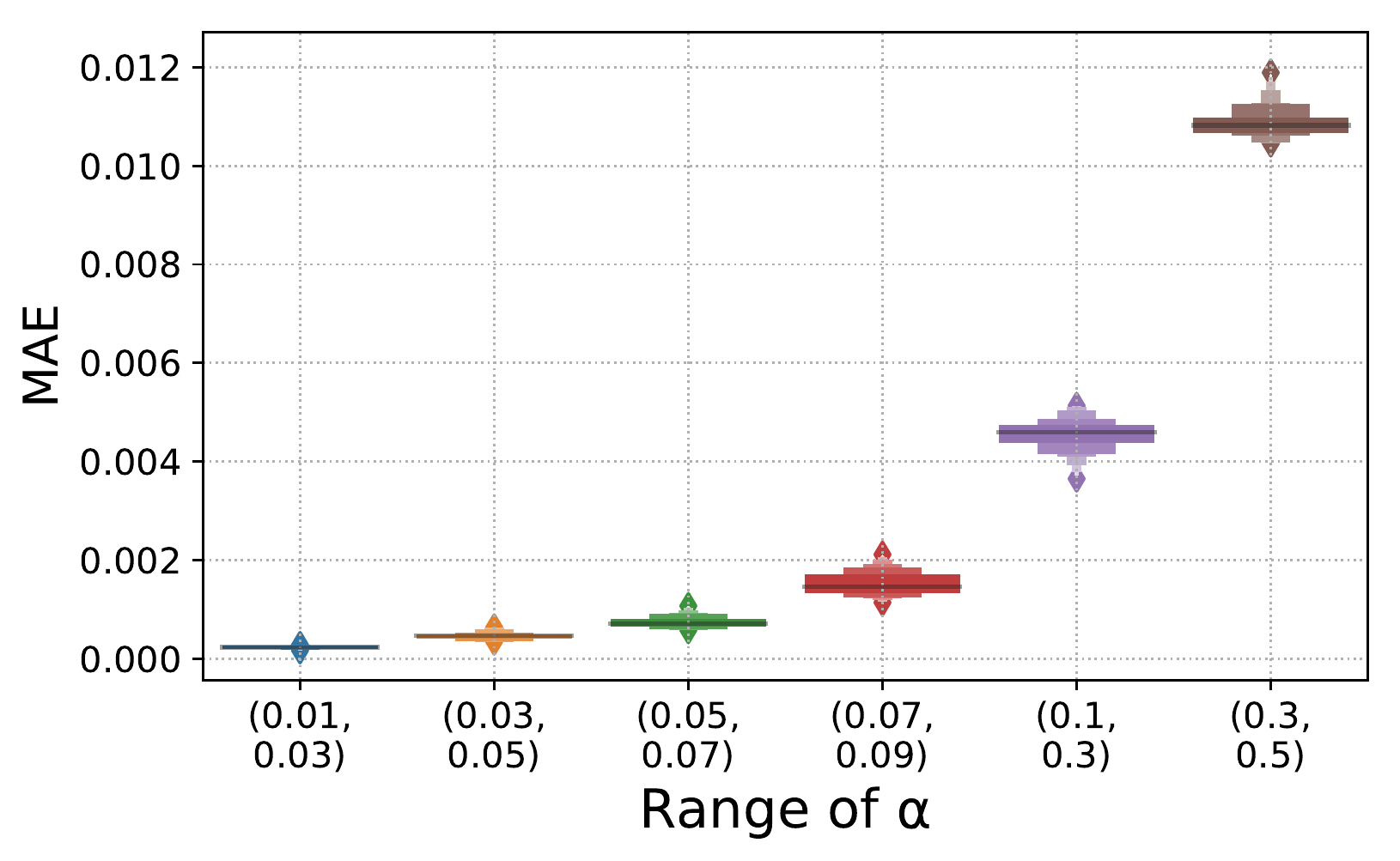}
    \caption{Mean absolute error of the estimated causal strength in different range of $\alpha$.}
    \label{fig:EM_param}
\end{figure}

\paragraph{Sensitivity Analysis}
In the sensitivity study, the experimental parameters are fixed with default settings mentioned in paragraph (b) while the control parameter will be traversed.
For the sensitivity of sample size, Fig. \ref{f2:c} shows that a large enough sample size is needed for all methods. In THP, $6000$ sample size is enough to achieve relatively good performance while other baselines require at least $15000$ or more. For the rest parameters, THP is relatively not sensitive compared with the baselines, which shows the robustness of our method. On the contrary, the baselines are sensitive to either the number of event types or the range of $\alpha$ and $\mu$. In particular, THP\_S is not sensitive in most cases, which shows the necessity of the topological information. 

Figure \ref{fig:EM_param} shows the Mean Absolute Error (MAE) between the estimated and the ground truth causal strengths with different ranges of $\alpha$, in which $MAE=\frac{1}{K|\mathbf{V} |^{2}}\sum _{v\in \mathbf{V}}\sum _{v'\in \mathbf{V}}\sum _{k=0}^{K} |\alpha _{v',v,k} -\hat{\alpha }_{v',v,k} |$ where $\hat{\alpha }_{v',v,k}$ denotes the ground truth causal strength.
We can see that we have an overall small MAE (up to 0.012) even though the scale of $\alpha$ goes to $[0.3, 0.5]$, which shows that the proposed method can indeed obtain the true causal strength with a relatively small error and verifies the effectiveness of the proposed method.

Table \ref{tab:gaussian} and Table \ref{tab:uniform} show the sensitive analysis of different kernel types. Interestingly, even though the types of kernels are different from our implemented exponential kernel, THP still achieves the best result in all cases, which shows the robustness of our method.

\begin{table}[tb]
	\centering

	\caption{Results on Real World Data }
	\begin{tabular}{l|rrr}
		\toprule
		& \textbf{Recall} & \textbf{Precision} & \textbf{F1} \\
		\midrule
		\textbf{PCMCI} & 0.30851 & 0.46774 & 0.37179 \\
		\textbf{NPHC} & 0.34042 & 0.38554 & 0.36158 \\
		\textbf{MLE-SGL} & 0.21276 & 0.47619 & 0.29411 \\
		\textbf{ADM4} & 0.34042 & 0.49231 & 0.40252 \\
		\textbf{THP-NT} & 0.36170 & 0.56667 & 0.44156 \\
		\textbf{THP-S} & 0.39362 & 0.49333 & 0.43787 \\
		\textbf{THP} &\textbf{0.39362} & \textbf{0.72549} & \textbf{0.51034} \\
		\bottomrule
	\end{tabular}
	
	\label{tab:1}

\end{table}

\begin{figure}[t]
	\centering
	\subfigure[F1 in the Real Data Experiments]{
		\includegraphics[width=0.45\textwidth]{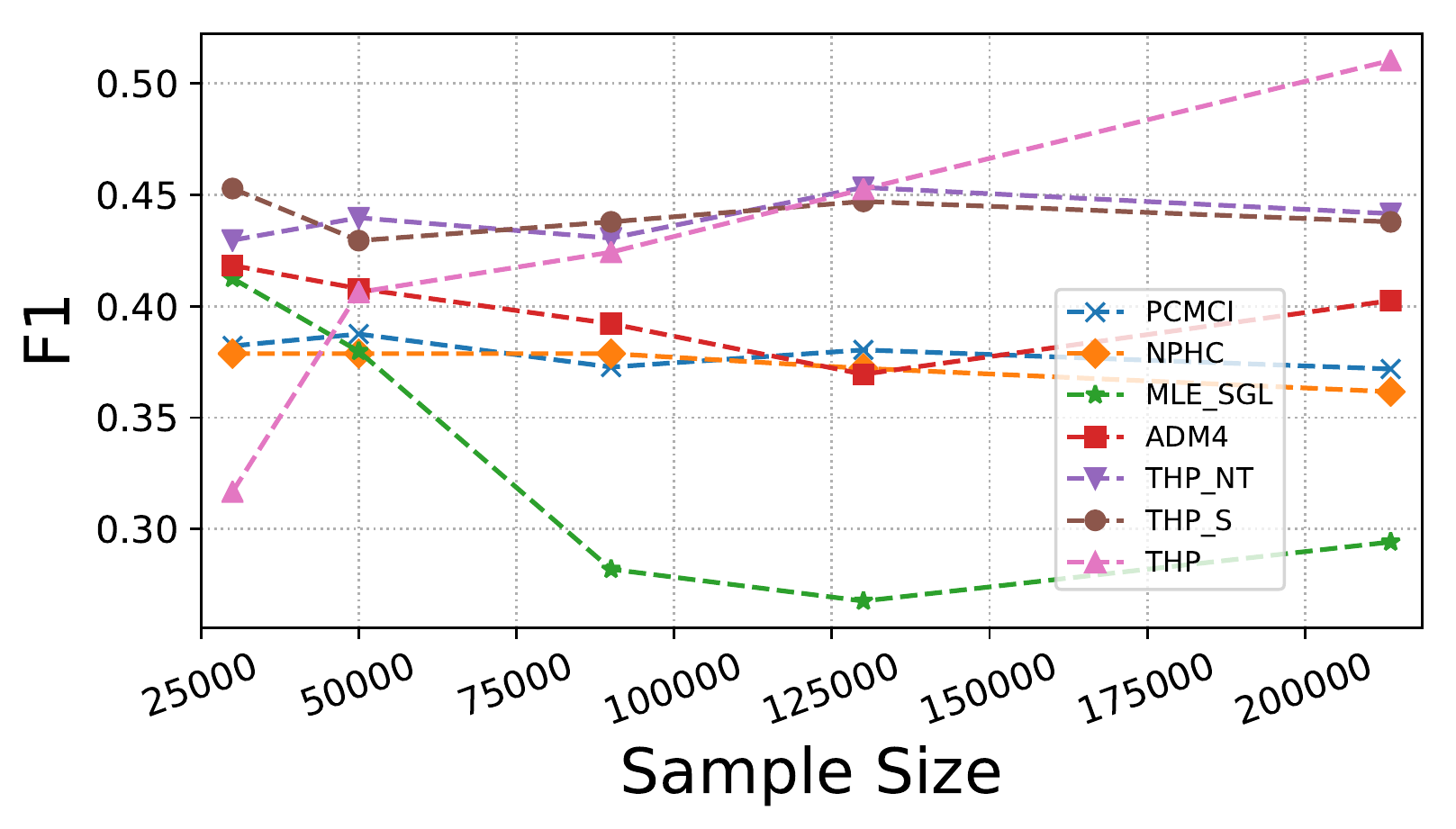}
	}
	\subfigure[Precision in the Real Data Experiments]{
	\includegraphics[width=0.45\textwidth]{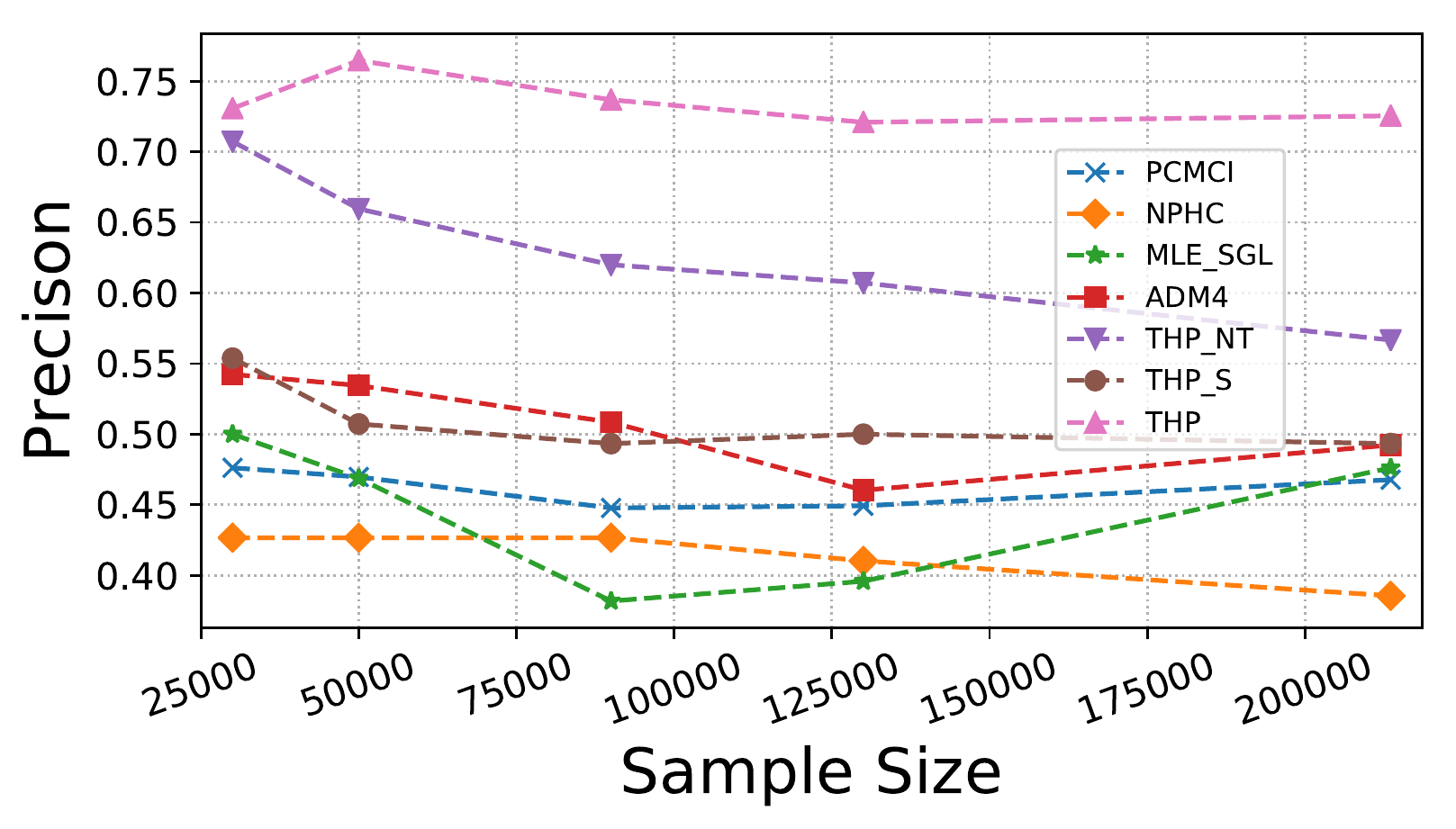}
}
	\subfigure[Recall in the Real Data Experiments]{
	\includegraphics[width=0.45\textwidth]{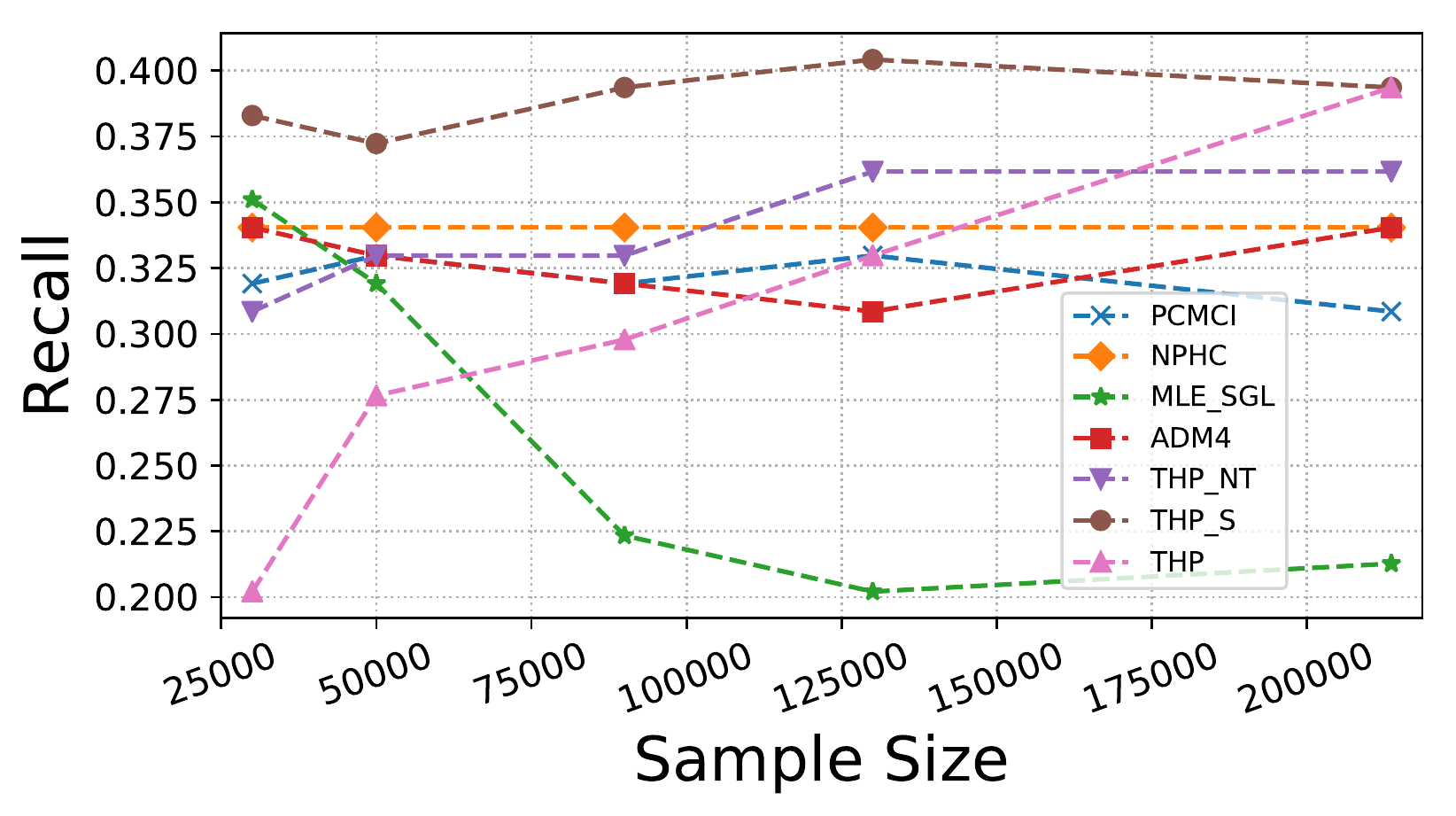}
}
	\caption{ Real Data Experiments in Different Sample Size}	
	\label{fig:diff_event_len}
\end{figure}

\subsection{Real World Data}
\paragraph{Data Description and Processing}
In this part, we test our algorithms on a very challenging dataset from a real metropolitan cellular network, to find causal structure among 18 types of alarms. It has 3087 network elements, the minimal occurrence times of the alarms are 2000, and there is a total of 228030 alarms in the data set. In this dataset, we regard the network elements as nodes and the physical connections between network elements as edges in the topological graph $\mathcal{G}_N$ in which alarms can only propagate between physically connected network elements. 
Note that, the causal impact mechanism of many alarms is not deterministic but probabilistic. For example, a high-temperature alarm may be the reason for the increased probability of other alarms occurring, but it does not necessarily cause other alarms to occur. 
Moreover, there are deviations between the real occurrence and recording of alarms, which makes it a more challenging dataset. The description of the real-world data is provided in Appendix E.

\paragraph{Comparison with Baselines}
As shown in Table \ref{tab:1}, our method outperforms the compared methods. However, in terms of recall, the performances of all methods are not well enough and the possible reason is that the period of the data is only one week and some causal relationships may have relatively weak causal strength which will not be detected. In fact, how to deal with the weak causal strength is still an open problem and we will leave it as future work. As for precision, THP shows a relatively high precision which is important for real-world applications.

\paragraph{Sensitivity of Sample Size} By choosing different starting points in the time span, we tested our method in different sample sizes. The results are given in Figure \ref{fig:diff_event_len}, which shows the importance of the sample size. It is also interesting to see that the number of samples has a higher impact on the recall than precision as the precision remains around 0.7 in all starting points while the recall increases as the sample size grows. The reason is that in the real world data set, the causal events tend to occur in different time spans such that we will need a large enough time span to contain all the causal events.

\paragraph{Case Study}
we successfully inferred some causal relationships that can spread throughout the network elements and verified them against those provided by domain experts. 
We visualize the causal strength at different topology distances in Fig. \ref{f5}. As shown in the figure, the causal strength decreases as the distance $k$ increases, which is consistent with our common sense that an event is usually influenced by its neighbors. Another important finding is that some edges only have strong causal strength at $k=1$ (while not at $k=0$), such as NCB$\rightarrow$NNL and ELD$\rightarrow$SSC. It implies that these kinds of event types are only triggered by their neighbors. These kinds of causality will be missed if the topological structure is ignored. The operation and maintenance manual also supports our findings. For example, consider ELD$\rightarrow$SSC, where SSC indicates that synchronous is changed in S1 mode and ELD indicates that the link connected to an Ethernet port is down. According to the operation and maintenance manual, SSC may be caused by fault of fiber connection of the alarm, such as ELD, generated by the upstream network elements which are the neighbors.

\section{Conclusion}
In this work, we propose a topological Hawkes process for learning causal structure on event sequences. By extending the convolution in the time domain to the joint convolution in the graph-time domain, the model takes both the topological constraint and the causal constraint in a unified likelihood framework and successfully recovers the causal structure among the event types. To the best of our knowledge, this is the first causal structure learning method for the event sequences with topological constraints. The success of THP not only provides an effective solution for the real-world event sequences but also shows a promising direction for the causal discovery on the non-i.i.d. samples. In the future, we plan to extend our work to a general point process with a more general graph convolutional kernel.

\appendices

\setcounter{theorem}{0}
\setcounter{remark}{0}
\setcounter{corollary}{0}

\setcounter{equation}{0}
\numberwithin{equation}{section}

\setcounter{figure}{0}

\setcounter{section}{0}

\section{Derivation of Intensity Function}
Recall Eq. (5) and Eq. (6) in the main article, the intensity function of THP is written as follows:
\begin{gather}
  \label{eq:appendix_convolution_time}
s_{v',v,t}=( \phi _{v',v} *dC_{v'})_{\mathbf{T}}( t),\\
  \lambda _{v} (n,t)=\mu _{v} +\sum _{v'\in \mathbf{V}}( g_{v',v} *s_{v',v,t})_{\mathcal{G}_{N}} (n).
\label{eq:appendix_convolution_graph}
\end{gather}
By introducing the graph convolution in Section 3.2, Eq. \ref{eq:appendix_convolution_graph} can be rewritten as:
\begin{equation}
\lambda _{v} (n,t)=\mu _{v} +\sum _{v'\in \mathbf{V}}(g_{v',v}( L ) s_{v',v,t})(n).
\end{equation}
Let the graph convolution kernel be $g_{v',v} (L )=\sum ^{K}_{k=0} \theta_{v',v,k} (I-L )^{k}$ and temporal convolution be $\phi_{v',v}=\beta_{v',v}\kappa(t)$, the intensity function can be reformulated as
\begin{equation}
\small
\begin{aligned}
 & \lambda _{v} (n,t)\\
= & \mu _{v} +\sum _{v'\in \mathbf{V}}\sum _{k=0}^{K} \theta _{v',v,k}\left( (I-L )^{k} s_{v',v,t}\right) (n)\\
= & \mu _{v} +\sum _{v'\in \mathbf{V}}\sum _{k=0}^{K} \theta _{v',v,k}\left(\hat{A}^{k} s_{v',v,t}\right) (n)\\
= & \mu _{v} +\sum _{v'\in \mathbf{V}}\sum _{k=0}^{K} \theta _{v',v,k}\left(\hat{A}_{n,\cdot }^{k} s_{v',v,t}\right)\\
= & \mu _{v} +\sum _{v'\in \mathbf{V}}\sum _{k=0}^{K}\sum _{n'\in \mathbf{N}} \theta _{v',v,k}\hat{A}_{n',n}^{k}\int _{t\in \mathbf{T}_{t^{-}}} \beta _{v',v} \kappa (t-t')dC_{v} (n',t'),\\
= & \mu _{v} +\sum _{v'\in \mathbf{V}}\sum _{k=0}^{K}\sum _{n'\in \mathbf{N}} \alpha _{v',v,k}\hat{A}_{n',n}^{k}\int _{t\in \mathbf{T}_{t^{-}}} \kappa (t-t')dC_{v} (n',t'),
\end{aligned}
\end{equation}
where $\hat{A}^k_{n',n}$ is the $n',n$ entry of matrix $\hat{A}^k$ and $\hat{A}^{k}_{n,\cdot }\in \mathbb{R}^{1\times |N|}$ is the $n^{th}$ row vector of the matrix $\hat{A}^{k}$ and $\alpha _{v',v,k} =\theta _{v',v,k} \beta _{v',v}$.

\section{Proof of Remark \ref{thm:thm2}}

\begin{proof}
\noindent$\Longrightarrow$:
If $v\rightarrow v' \notin \mathbf{E}_V$, then according to Definition \ref{def:local_ind_topo}, there must exist a subset $S \subseteq \mathbf{V}$, such that $v \not\to v' |_{\mathcal{N}}S$. That is $v,v'$ will not directly connect otherwise we can not find such a $S$, and we must have $\alpha _{v',v,k}\hat{A}_{n',n}^{k}= 0$ for all $n,n' \in \mathbf{N}$, $k\in \{0,...,K\}$.

\noindent$\Longleftarrow $:
Suppose that $v\to v'\in \mathbf{E}_V$ then for any subset $S \subseteq \mathbf{V}$, we have $v \to v' |_{\mathcal{N}}S$. Therefore, $v,v'$ should be directly connected and there exists $k\in \{0,...,K\}$, $n,n'\in \mathbf{N}$ such that $\alpha _{v',v,k}\hat{A}_{n',n}^{k}\ne 0$.
\end{proof}

\section{Derivation of Likelihood Function}
\label{appendix:Likelihood}
In the discrete time domain, assuming that the observed event sequences $\mathbf{X}$, the likelihood function can be derived as follows:

\begin{equation}
\label{L_in_appendix}
\begin{aligned}
 & L(\mathcal{G}_{V} ,\Theta ;\mathbf{X} ,\mathcal{G}_{N} )\\
= & \sum\limits _{v\in \mathbf{V}}\sum\limits _{t\in \mathbf{T}}\sum _{n\in \mathbf{N}} P\left( X_{n,v,t} |H_{t}^{\mathbf{PA}_{v}}\right)\\
= & \sum\limits _{v\in \mathbf{V}}\sum\limits _{t\in \mathbf{T}}\sum _{n\in \mathbf{N}}\log\left[\frac{e^{-\lambda _{v} (n,t))\Delta t}}{X_{n,v,t} !}  (\lambda _{v} (n,t)\Delta t)^{X_{n,v,t}}\right]\\
= & \sum\limits _{v\in \mathbf{V}}\sum\limits _{t\in \mathbf{T}}\sum _{n\in \mathbf{N}} [-\lambda _{v} (n,t)\Delta t+X_{n,v,t}\log (\lambda _{v} (n,t))]\\
 & +\underbrace{\sum _{v\in \mathbf{V}}\sum _{t\in \mathbf{T}}\sum _{n\in \mathbf{N}} [-\log (X_{n,v,t} !)+X_{n,v,t}\log (\Delta t)]}_{Const}\\
 & \\
 & \lambda _{v} (n,t)\\
= & \mu _{v} +\sum _{v'\in \mathbf{PA}_{v}}\sum _{n'\in \mathbf{N}}\sum _{k=0}^{K} \alpha _{v',v,k}\hat{A}_{n',n}^{k}\sum _{t'\in \mathbf{T}_{t^{-}}} \kappa (t-t')X_{n',v',t'}
\end{aligned}
\end{equation}

\section{Derivation of EM Algorithm}
\label{EM}

Following the work \cite{lewis2011nonparametric,zhou2013learning}, we utilize the EM algorithm to search the optimal parameters $\Theta$. We first introduce hidden variables $z_{n,v,t}$, and let $z_{n,v,t} =(n'\rightarrow n,v'\rightarrow v,t'\rightarrow t,k)$
indicate that an event with type $v$ in node $n$ at time $t$ is triggered by type $v'$ events in node $n'$ at $t'$, through a path of length $k$. When $z_{n,v,t} =(n\rightarrow n,v\rightarrow v,t\rightarrow t,0)$, it means that the event occurs spontaneously. 
By modeling such latent variable, the EM algorithm for $\Theta$ consists of two steps. First is the expectation step, which construct the lower bound $Q({\Theta } ,{\Theta }^{( i-1)})$ by given the last estimation parameters ${\Theta }^{( i-1)}$. Second is the maximization step, which maximum $Q$ to obtain the best estimation in $i$ step ${\Theta }^{( i)}= \arg\max_{\Theta} Q({\Theta } ,{\Theta }^{( i-1)})$.

Generally, we first randomly initialize the parameters $\Theta^{(0)}$, then perform the following E-M steps iteratively until $Q$ converges. The derivation of the updated rule in E-step and M-step are given as follows:
\subsection*{Expectation step:}
In the E step, we construct the lower bound for the last parameter estimation ${\Theta }^{( i-1)}$ as follows:

\begin{equation}
\small
\label{Q}  
\begin{aligned}
 & Q\left( \Theta ,\Theta ^{(i-1)}\right)\\
= & \sum\limits _{n\in \mathbf{N}}\sum _{v\in \mathbf{V}}\sum\limits _{t\in \mathbf{T}}\biggl[\biggl(\sum _{n'\in \mathbf{Ne}_{n}}\sum _{v'\in \mathbf{PA}_{v}}\sum _{t'\in \mathbf{T}_{t^{-}}}\sum _{k=0}^{K} q_{n,v,t}^{\alpha } (n',v',t',k)\\
 & \times \log p\left( X_{n,v,t} ,z_{n,v,t} =\left( n'\rightarrow n,v'\rightarrow v,t'\rightarrow t,k\right) ,H_{t}^{\mathbf{PA}_{v}} ,\Theta \right)\biggr)\\
 & +q_{n,v,t}^{\mu }\log p\left( X_{n,v,t} ,z_{n,v,t} =\left( n\rightarrow n,v\rightarrow v,t\rightarrow t,0\right) ,H_{t}^{\mathbf{PA}_{v}} ,\Theta \right)\biggr]\\
= & \sum\limits _{n\in \mathbf{N}}\sum _{v\in \mathbf{V}}\sum\limits _{t\in \mathbf{T}}\biggl[\sum _{n'\in \mathbf{Ne}_{n}}\sum _{v'\in \mathbf{PA}_{v}}\sum _{t'\in \mathbf{T}_{t^{-}}}\sum _{k=0}^{K} q_{n,v,t}^{\alpha } (n',v',t',k)\\
 & \times X_{n,v,t}\log (\alpha _{v',v,k}\hat{A}_{n',n}^{k} \kappa (t-t')X_{n',v',t'} )\biggr]\\
 & +\sum\limits _{n\in \mathbf{N}}\sum _{v\in \mathbf{V}}\sum\limits _{t\in \mathbf{T}} q_{n,v,t}^{\mu } X_{n,v,t}\log \mu _{v}\\
 & -\sum\limits _{n\in \mathbf{N}}\sum _{v\in \mathbf{V}}\sum\limits _{t\in \mathbf{T}} \lambda _{v} (n,t)\Delta t+Const,
\end{aligned}
\end{equation}
where $p\left( X_{n,v,t} ,z_{n,v,t} =\left( n'\rightarrow n,v'\rightarrow v,t'\rightarrow t,k\right) ,H_{t}^{\mathbf{PA}_{v}} ,\Theta \right) =\frac{e^{-\lambda _{v} (n,t))\Delta t}}{X_{n,v,t} !} (\alpha _{v',v,k} A_{n',n}^{k} \kappa (t-t')X_{n',v',t'} \Delta t)^{X_{n,v,t}}$ is the likelihood function at $\displaystyle k\neq 0$. For the case that $\displaystyle k=0$, we have $p\left( X_{n,v,t} ,z_{n,v,t} =\left( n\rightarrow n,v\rightarrow v,t\rightarrow t,0\right) ,H_{t}^{\mathbf{PA}_{v}} ,\Theta \right) =\frac{e^{-\lambda _{v} (n,t))\Delta t}}{X_{n,v,t} !} (\mu _{v} \Delta t)^{X_{n,v,t}}$.

Moreover, $q_{n,v,t}^{\mu } \\=p\left( z_{n,v,t} =\left( n\rightarrow n,v\rightarrow v,t\rightarrow t,0\right) |X_{n,v,t} ,H_{t}^{\mathbf{PA}_{v}} ,\Theta ^{(i-1)}\right)$ indicates the probability of event $v$ occurred spontaneously in node $n$ at $t$, i.e.,
\begin{equation}
\small
\begin{aligned}
 & q_{n,v,t}^{\mu }\\
= & p\left( z_{n,v,t} =\left( n\rightarrow n,v\rightarrow v,t\rightarrow t,0\right) |X_{n,v,t} ,H_{t}^{\mathbf{PA}_{v}} ,\Theta ^{(i-1)}\right)\\
= & \frac{\mu _{v}^{(i-1)}}{\lambda _{v}^{(i-1)} (n,t)} ,
\end{aligned}
\end{equation}
where
\begin{equation}
\small
\begin{aligned}
 & \lambda _{v}^{(i-1)} (n,t)\\
= & \mu _{v}^{(i-1)} +\sum _{v'\in \mathbf{PA}_{v}}\sum _{n'\in \mathbf{N}}\sum _{k=0}^{K} \alpha _{v',v,k}^{(i-1)}\hat{A}_{n',n}^{k}\sum _{t'\in \mathbf{T}_{t^{-}}} \kappa (t-t')X_{n',v',t'} ,
\end{aligned}
\end{equation}
while 
$ 
q_{n,v,t}^{\alpha } (n',v',t',k)\\
=p\left( z_{n,v,t} =\left( n'\rightarrow n,v'\rightarrow v,t'\rightarrow t,k\right) |X_{n,v,t} ,H_{t}^{\mathbf{PA}_{v}} ,\Theta ^{(i-1)}\right)
$ indicates the probability of event $(n,v,t)$ is triggered by event $(n',v',t')$ through a path of length k, and we have
\begin{equation}
\small
\begin{aligned}
 & q_{n,v,t}^{\alpha } (n',v',t',k)\\
= & p\left( z_{n,v,t} =\left( n'\rightarrow n,v'\rightarrow v,t'\rightarrow t,k\right) |X_{n,v,t} ,H_{t}^{\mathbf{PA}_{v}} ,\Theta ^{(i-1)}\right)\\
= & \frac{\alpha _{v',v,k}^{(i-1)}\hat{A}_{n',n}^{k} \kappa (t-t')X_{n',v',t'}}{\lambda _{v}^{(i-1)} (n,t)} .
\end{aligned}
\end{equation}

\subsection*{Maximization step: }
In M step, we aim to find a $\Theta^{(i)}$ that maximizes $Q\left({\Theta^{(i)} } ,{\Theta }^{{(i)} -1}\right)$. By using the KKT condition:
\begin{equation}
\begin{cases}
\frac{\partial Q\left( \Theta ^{(i)} ,\Theta ^{(i-1)}\right)}{\partial \mu ^{(i)}_{v}} & =0\\
\\
\frac{\partial Q\left( \Theta ^{(i)} ,\Theta ^{(i-1)}\right)}{\partial \alpha ^{(i)}_{v',v,k}} & =0,
\end{cases}
\end{equation}
we obtain the close form solution of the maximum values:
\begin{equation}
\small
\label{appendix_update mu} 
\begin{aligned}
\mu ^{(i)}_{v} & =\frac{\sum\limits _{n\in \mathbf{N}}\sum\limits _{t\in \mathbf{T}} q^{\mu }_{n,v,t} X_{n,v,t}}{|\mathbf{N} ||\mathbf{T} |\ \Delta t} ,
\end{aligned}
\end{equation}

\begin{equation}
\small
\label{appendix_update alpha} 
\begin{aligned}
\alpha ^{(i)}_{v',v,k} & =\frac{\sum\limits _{n\in \mathbf{N}}\sum\limits _{t\in \mathbf{T}}\left[\sum _{n'\in \mathbf{N}}\sum _{t'\in \mathbf{T}_{t^{-}}} q^{\alpha }_{n,v,t} (n',v',t',k)\right] X_{n,v,t}}{\sum _{n\in \mathbf{N}}\sum\limits _{t\in \mathbf{T}}\left[\sum _{n'\in \mathbf{N}}\sum _{t'\in \mathbf{T}_{t^{-}}} \hat{A}^{k}_{n',n} \kappa ( t-t') X_{n',v',t'}\right] \Delta t}.
\end{aligned}
\end{equation}

\begin{table*}[t]
  \centering
  \caption{Ground True}
    \begin{tabular}{cc|cc}
    \toprule
    Cause & Effect & Cause & Effect \\
    \midrule
    MW\_RDI & LTI   & MW\_BER\_SD & LTI \\
    MW\_RDI & CLK\_NO\_TRACE\_MODE & MW\_BER\_SD & S1\_SYN\_CHANGE \\
    MW\_RDI & S1\_SYN\_CHANGE & MW\_BER\_SD & PLA\_MEMBER\_DOWN \\
    MW\_RDI & LAG\_MEMBER\_DOWN & MW\_BER\_SD & MW\_RDI \\
    MW\_RDI & PLA\_MEMBER\_DOWN & MW\_BER\_SD & MW\_LOF \\
    MW\_RDI & ETH\_LOS & MW\_BER\_SD & ETH\_LINK\_DOWN \\
    MW\_RDI & ETH\_LINK\_DOWN & MW\_BER\_SD & NE\_COMMU\_BREAK \\
    MW\_RDI & NE\_COMMU\_BREAK & MW\_BER\_SD & R\_LOF \\
    MW\_RDI & R\_LOF & R\_LOF & LTI \\
    TU\_AIS & LTI   & R\_LOF & S1\_SYN\_CHANGE \\
    TU\_AIS & CLK\_NO\_TRACE\_MODE & R\_LOF & LAG\_MEMBER\_DOWN \\
    TU\_AIS & S1\_SYN\_CHANGE & R\_LOF & PLA\_MEMBER\_DOWN \\
    RADIO\_RSL\_LOW & LTI   & R\_LOF & ETH\_LINK\_DOWN \\
    RADIO\_RSL\_LOW & S1\_SYN\_CHANGE & R\_LOF & NE\_COMMU\_BREAK \\
    RADIO\_RSL\_LOW & LAG\_MEMBER\_DOWN & LTI   & CLK\_NO\_TRACE\_MODE \\
    RADIO\_RSL\_LOW & PLA\_MEMBER\_DOWN & HARD\_BAD & LTI \\
    RADIO\_RSL\_LOW & MW\_RDI & HARD\_BAD & CLK\_NO\_TRACE\_MODE \\
    RADIO\_RSL\_LOW & MW\_LOF & HARD\_BAD & S1\_SYN\_CHANGE \\
    RADIO\_RSL\_LOW & MW\_BER\_SD & HARD\_BAD & BD\_STATUS \\
    RADIO\_RSL\_LOW & ETH\_LINK\_DOWN & HARD\_BAD & POWER\_ALM \\
    RADIO\_RSL\_LOW & NE\_COMMU\_BREAK & HARD\_BAD & LAG\_MEMBER\_DOWN \\
    RADIO\_RSL\_LOW & R\_LOF & HARD\_BAD & PLA\_MEMBER\_DOWN \\
    BD\_STATUS & S1\_SYN\_CHANGE & HARD\_BAD & ETH\_LOS \\
    BD\_STATUS & LAG\_MEMBER\_DOWN & HARD\_BAD & MW\_RDI \\
    BD\_STATUS & PLA\_MEMBER\_DOWN & HARD\_BAD & MW\_LOF \\
    BD\_STATUS & ETH\_LOS & HARD\_BAD & ETH\_LINK\_DOWN \\
    BD\_STATUS & MW\_RDI & HARD\_BAD & NE\_COMMU\_BREAK \\
    BD\_STATUS & MW\_LOF & HARD\_BAD & R\_LOF \\
    BD\_STATUS & ETH\_LINK\_DOWN & HARD\_BAD & NE\_NOT\_LOGIN \\
    BD\_STATUS & RADIO\_RSL\_LOW & HARD\_BAD & RADIO\_RSL\_LOW \\
    BD\_STATUS & TU\_AIS & HARD\_BAD & TU\_AIS \\
    NE\_COMMU\_BREAK & LTI   & ETH\_LOS & LTI \\
    NE\_COMMU\_BREAK & CLK\_NO\_TRACE\_MODE & ETH\_LOS & CLK\_NO\_TRACE\_MODE \\
    NE\_COMMU\_BREAK & S1\_SYN\_CHANGE & ETH\_LOS & S1\_SYN\_CHANGE \\
    NE\_COMMU\_BREAK & LAG\_MEMBER\_DOWN & ETH\_LOS & LAG\_MEMBER\_DOWN \\
    NE\_COMMU\_BREAK & PLA\_MEMBER\_DOWN & ETH\_LOS & PLA\_MEMBER\_DOWN \\
    NE\_COMMU\_BREAK & ETH\_LOS & ETH\_LOS & ETH\_LINK\_DOWN \\
    NE\_COMMU\_BREAK & ETH\_LINK\_DOWN & MW\_LOF & LTI \\
    NE\_COMMU\_BREAK & NE\_NOT\_LOGIN & MW\_LOF & CLK\_NO\_TRACE\_MODE \\
    ETH\_LINK\_DOWN & LTI   & MW\_LOF & S1\_SYN\_CHANGE \\
    ETH\_LINK\_DOWN & CLK\_NO\_TRACE\_MODE & MW\_LOF & LAG\_MEMBER\_DOWN \\
    ETH\_LINK\_DOWN & S1\_SYN\_CHANGE & MW\_LOF & PLA\_MEMBER\_DOWN \\
    S1\_SYN\_CHANGE & LTI   & MW\_LOF & ETH\_LOS \\
    POWER\_ALM & BD\_STATUS & MW\_LOF & MW\_RDI \\
    POWER\_ALM & ETH\_LOS & MW\_LOF & ETH\_LINK\_DOWN \\
    POWER\_ALM & MW\_RDI & MW\_LOF & NE\_COMMU\_BREAK \\
    POWER\_ALM & MW\_LOF & MW\_LOF & R\_LOF \\
    \bottomrule
    \end{tabular}
  \label{tab:ground_truth}
\end{table*}

\section{Real-World Data Description}

The real-world data used in this paper are the collection of the alarms records that occurred in a metropolitan cellular network within a week. This data set includes an alarm table and a topological table of network elements. 
We will release the data set after this work is accepted.

\subsection{Alarm Table}

The alarm table consists of 4655592 records, containing 5 fields including 'Alarm Name', 'First Occurrence', 'Cleared On', 'Last Occurrence', and 'Alarm Source'. The meaning of the fields is as follows:

\begin{itemize}
\item 'Alarm Name': The name of alarm type;
\item 'First Occurrence': The time when the alarm first occurred;

\item 'Last Occurrence' The time when the alarm last occurred;

\item 'Cleared On': The time when the alarm is cleared;

\item 'Alarm Source' The id of the network element where the alarm occurred.
\end{itemize}

There are 172 types of alarms in the data set. Some typical alarms are as follows: 'ETH\_LOS' alarm indicates the loss of connection on an optical Ethernet port, 'MW\_LOF' alarm indicates loss of microwave frames and 'HARD\_BAD' alarm indicates that the hardware is faulty.

\subsection{Topological Table}

The topological table contains 4 fields including 'Path ID', 'NE\_NAME', 'NE\_TYPE', and 'PATH\_HOP'. The specific meanings of the fields are as follows: 
\begin{itemize}
\item 'Path ID': The id of the path in the metropolitan cellular network;
\item 'NE\_NAME': The name of the network element, which corresponds to the 'Alarm Source' in the alarm table;
\item 'NE\_TYPE': The type of network element in the path, including ROUTER, MICROWAVE, and NODEB;
\item 'PATH\_HOP': The relative position of the network element in the path.
\end{itemize}

Besides, there are 41143 network elements involved in the topological table.

\subsection{Ground Truth}
The ground truth of the causal relationships among alarm types are provided by domain experts. Thus, in our experiment, we only select the sample whose alarm type is labeled by expert. Moreover, since there are large number of network elements involved in this data set but the time span is small, some of the alarm types might suffer from data insufficiency which may lead to unreliable results. To avoid this problem, we further filter the alarm types with more than 2000 occurrences, such that 18 types of alarms and 3087 network elements are involved. The ground truth of the causal relationships among the alarm types are given in Table \ref{tab:ground_truth}.

\section{Details of Baseline Methods}
In this section, we briefly introduce each baseline method and its corresponding settings.
\subsection{THP\_NT \& THP\_S}
THP\_NT and THP\_S are two variations of THP. THP\_NT is a version of THP that do not consider the topological structure of nodes. The intensity function of THP\_NT is given in Eq. \ref{TTHP-nt}, where the impact function does not depend on nodes. The same with THP, THP\_NT use two steps based searching algorithm to learn the causal graph. 

\begin{equation}
\small
    \label{TTHP-nt}
\lambda _{v} (n,t)=\mu _{v} +\sum _{v'\in \mathbf{V}}\sum _{t'\in \mathbf{T}_{t^{-}}} \alpha _{v',v} \kappa ( t-t') X_{n,v',t'} ,
\end{equation}

THP\_S takes topology information into intensity function like THP but does not use the sparse optimization scheme. Similar to ADM4, THP\_S first initialize causal structure as a complete graph. Then an EM algorithm is applied to optimize the parameters with a $\ell 1$ norm regularization on $\alpha_{v',v,k}$.

\begin{equation}
\small
\lambda _{v} (n,t)=\mu _{v} +\sum _{v'\in \mathbf{V}}\sum _{n'\in \mathbf{N}}\sum ^{K}_{k=0} \alpha _{v',v,k}\sum _{t'\in \mathbf{T}_{t^{-}}} \hat{A}^{k}_{n',n} \kappa ( t-t') X_{n',v',t'} ,
\end{equation}

In all THP and its variations methods, the decay kernel function $\kappa(t)$ is set to the exponential form $\kappa(t)=\exp(-\delta(t))$. In real-world data experiments, according to the alarms table, multiple alarms usually occur within a short time period. Thus, to focus on the short term effects, the decay parameters $\delta$ is set to $0.11$. 

\subsection{PCMCI}

PCMCI \cite{runge2019detecting} is a causal discovery algorithm for time series data based on conditional independence test. It consists of two stages: 1) PC$_1$ conditional selection to remove the irrelevant conditions for each variable by iterative independence testing and 2) use the momentary conditional independence (MCI) test to test whether the causal direction holds. Using different types of conditional independence tests, this method can be applied to different types of time series data. Here, PCMCI is applied to discover the causal relationship in event sequences. 
In our experiments, we use the mutual information as the independence test for the discrete event type data and set the time interval to $5$ seconds with max lag up to $2$.

\subsection{ADM4}

As in the intensity function, ADM4 \cite{zhou2013learningsocial} uses $\alpha _{v',v} \beta \exp (-\beta ( t-t') )$ to represent the impact function $\phi_{v',v}$ capturing the causal strength $\alpha_{v',v}$ from $v'$ to $v$. To further constraint the low-rank structure of the causal graph, the Nuclear and $\ell 1$-norm regularization is used. In detail, the nuclear norm $||A||_{*}$ is the sum of the singular value $\sum ^{|\mathbf{V} |}_{i=1} \sigma _{i}$, where $A\in \mathbb{R}^{|V|\times|V|}$ is the matrix of $\alpha_{v',v}$, and $\ell 1$ norm of $A$ is $||A||_{1} =\sum _{v',v} |\alpha_{v',v} |$. In our experiments, following the original work, the coefficients of $\ell 1$ and nuclear regularization are both $500$ and the parameter $\beta$ of impact function is $0.1$.

\begin{equation}
\small
  \label{adm4}
\begin{aligned}
\lambda _{v} (t) & =\mu _{v} +\sum _{v'\in \mathbf{V}}\int ^{t}_{t_{0}} \phi _{v',v} (t-t')dC_{v'}( t')\\
 & =\mu _{v} +\sum _{v'\in \mathbf{V}}\int ^{t}_{t_{0}} \alpha _{v',v} \beta \exp (-\beta ( t-t') )dC_{v'}( t')
\end{aligned}
\end{equation}

\subsection{NPHC}
NPHC \cite{achab2017uncovering} a nonparametric method that can estimate the matrix of integrated kernels of a multivariate Hawkes process. This method relies on the matching of the integrated order 2 and order 3 empirical cumulants to recover the Granger causality matrix $G_V$.

\subsection{MLE-SGL}

In MLE-SGL \cite{xu2016learning}, the impact function $\phi_{v',v}$ in the intensity is represented via a linear combination of basis functions $\kappa_m$ (sinc or Gaussian function), which is shown in \ref{mle-sgl}. In this model, $\sum ^{M}_{m=1} \alpha ^{m}_{v',v}$ is used to represent the causal strength from $v'$ to $v$. Combining with MLE, the EM base algorithm is proposed to optimize parameters and learn the causal structure. For sparsity of the causal graph, the sparse-group-lasso is used. More specifically, the sparse regularizer is denoted as $||A||_{1} =\sum _{v',v,m} |\alpha ^{m}_{v',v} |$ and the form of group-lasso is $||A||_{1,2} =\sum _{v',v} ||\alpha _{v',v} ||_{2}$, where $\alpha _{v',v} =\left[ \alpha ^{1}_{v',v} ,\alpha ^{2}_{v',v} ,\dotsc ,\alpha ^{m}_{v',v}\right]$. In our experiments, the number of Gaussian basis functions is $10$ and the coefficients of group-lasso and sparse regularization are $250$ and $750$.
\begin{equation}
\small
    \label{mle-sgl}
        \begin{aligned}
        \lambda _{v} (t) & =\mu _{v} +\sum _{v'\in \mathbf{V}}\int ^{t}_{t_{0}} \phi _{v',v} (t-t')dC_{v'}( t')\\
         & =\mu _{v} +\sum _{v'\in \mathbf{V}}\int ^{t}_{t_{0}}\sum ^{M}_{m=1} \alpha ^{m}_{v',v} \kappa _{m} (t-t' )dC_{v'}( t')
        \end{aligned}
\end{equation}

\bibliographystyle{IEEEtran}
\bibliography{TNNLS}

\ifCLASSOPTIONcaptionsoff
  \newpage
\fi

\begin{IEEEbiography}[{\includegraphics[width=1in, height=1.25in, clip, keepaspectratio]{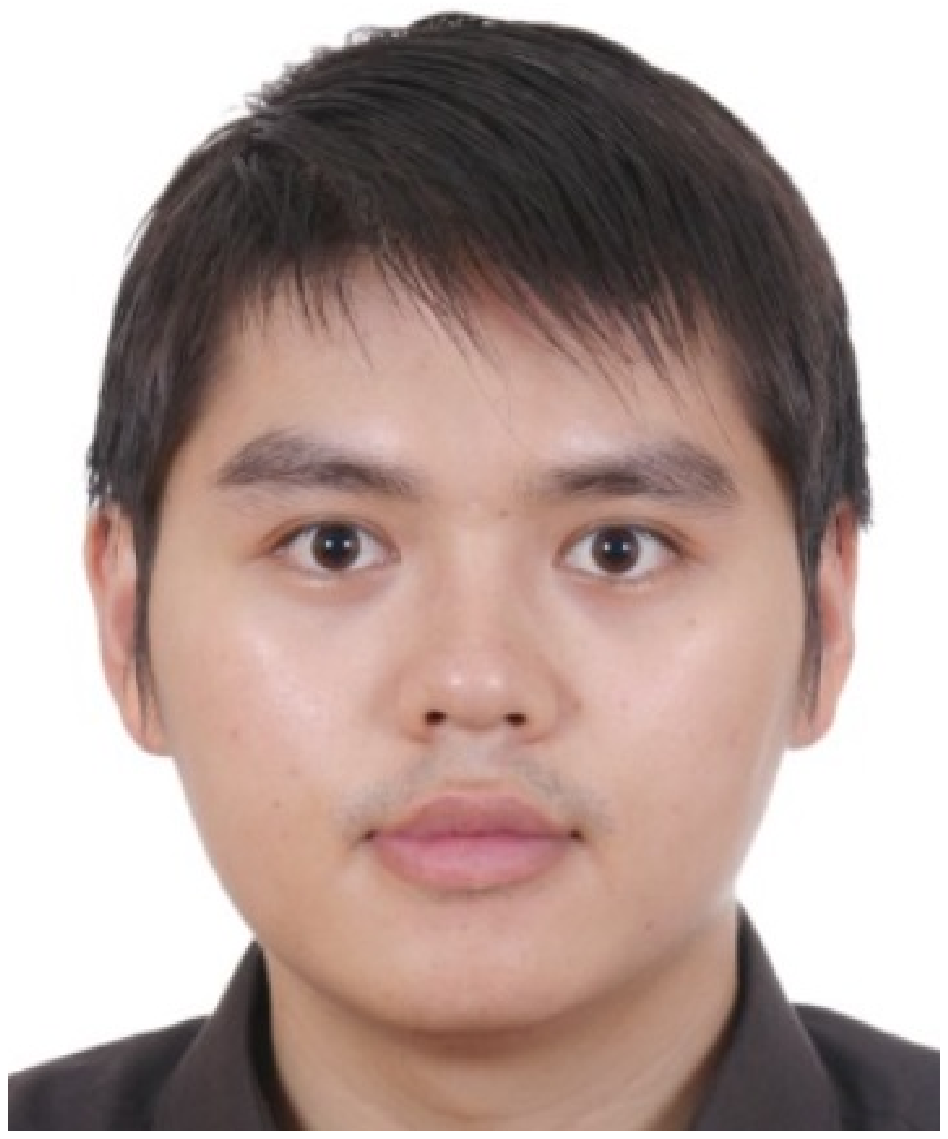}}]{Ruichu Cai} is currently a professor in the school of computer science and the director of the data mining and information retrieval laboratory, Guangdong University of Technology. He received his B.S. degree in applied mathematics and Ph.D. degree in computer science from South China University of Technology in 2005 and 2010, respectively. 
 
His research interests cover various topics, including causality, deep learning, and their applications. He was a recipient of the National Science Fund for Excellent Young Scholars, the Natural Science Award of Guangdong, and so on awards. He has served as the area chair of ICML 2022, NeurIPS 2022, and UAI 2022, senior PC for AAAI 2019-2022, IJCAI 2019-2022, and so on. He is now a senior member of CCF and IEEE.

\end{IEEEbiography}
\vspace{-8ex}

\begin{IEEEbiography}
[{\includegraphics[width=1in, height=1.25in, clip, keepaspectratio]{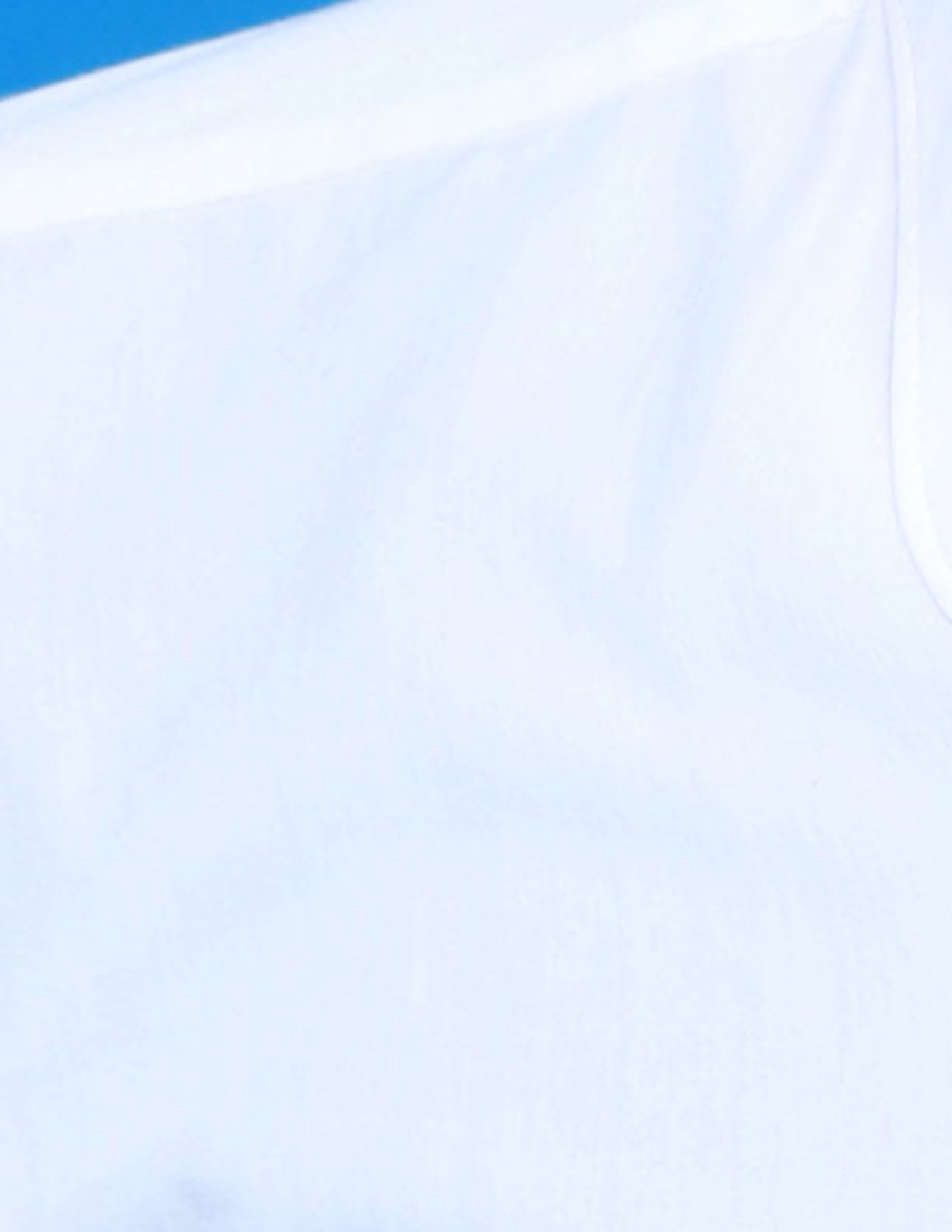}}]{Siyu Wu} received his B.E. degree in Computing Science 2019. He is currently pursuing the M.S. degree with the School of Computer, Guangdong University of Technology. 
\end{IEEEbiography}
\vspace{-8ex}

\begin{IEEEbiography}
[{\includegraphics[width=1in, height=1.25in, clip, keepaspectratio]{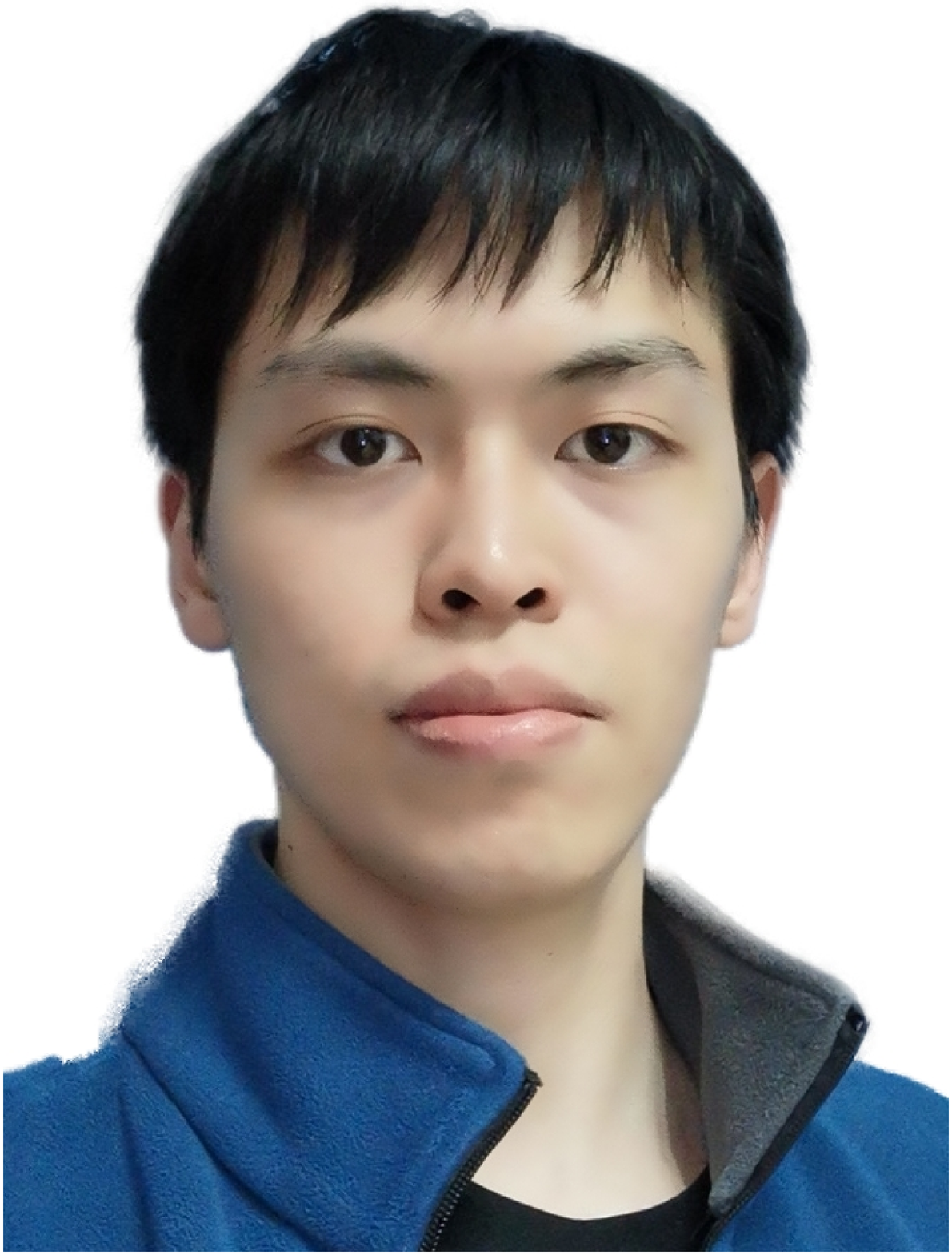}}]{Jie Qiao} received the Ph.D. from Guangdong University of Technology, school of computer science, in 2021.
His research interests include causal discovery and causality-inspired machine learning.
\end{IEEEbiography}

\vspace{-8ex}

\begin{IEEEbiography}
	[{\includegraphics[width=1in, height=1.25in, clip, keepaspectratio]{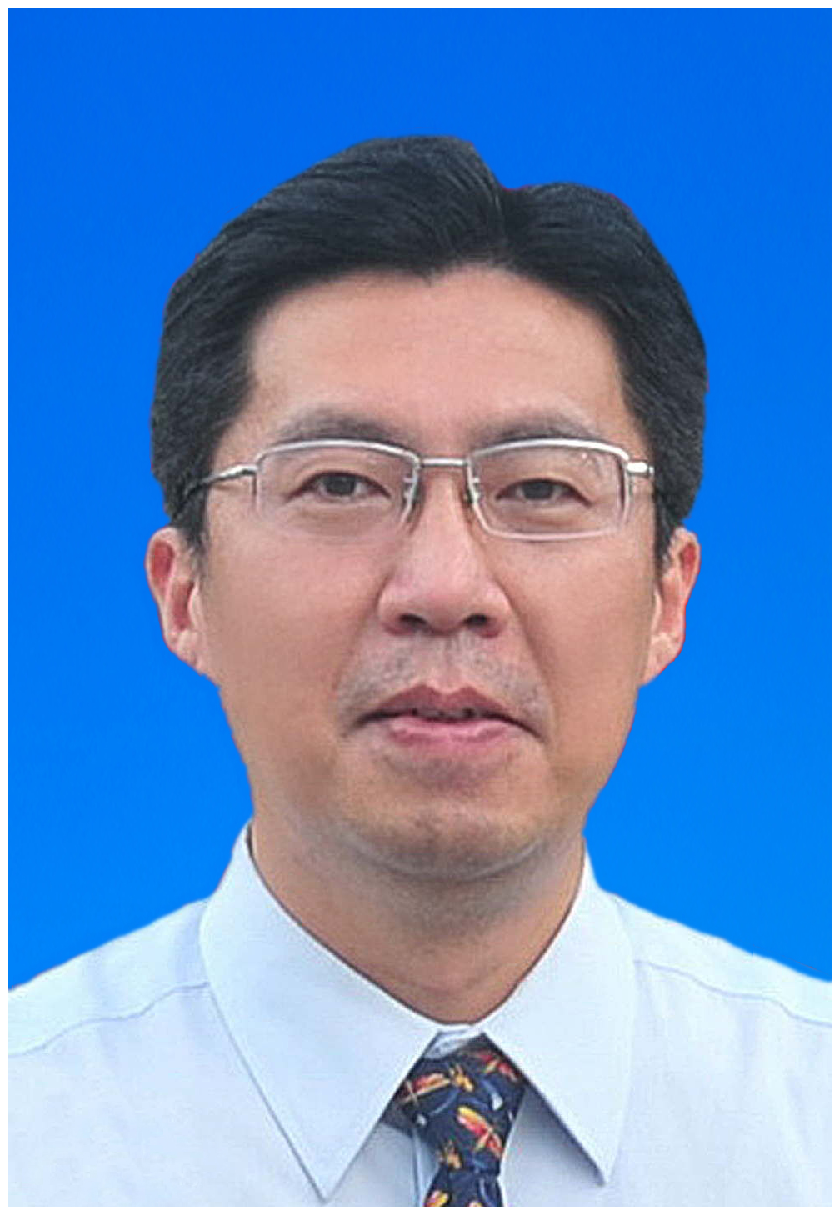}}]{Zhifeng Hao} received his B.S. degree in Mathematics from the Sun Yat-Sen University in 1990, and his Ph.D. degree in Mathematics from Nanjing University in 1995. He is currently a Professor in the School of Computer, Guangdong University of Technology, and College of Science, Shantou University. 
	
	His research interests involve various aspects of Algebra, Machine Learning, Data Mining, and Evolutionary Algorithms.
\end{IEEEbiography}

\vspace{-8ex}

\begin{IEEEbiography}
	[{\includegraphics[width=1in, height=1.25in, clip, keepaspectratio]{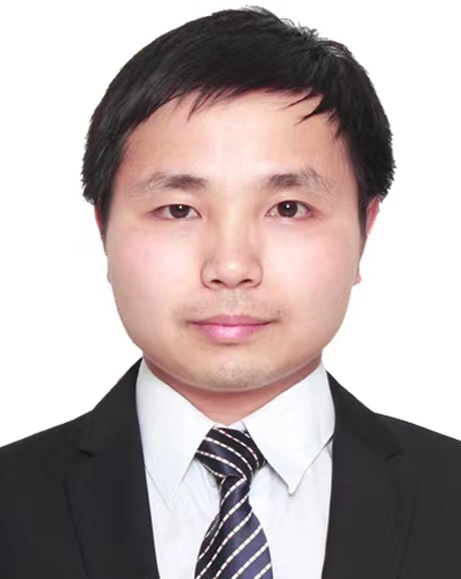}}]{Keli Zhang} received the MS. degree from Huazhong University of Science and Technology, Wuhan, China, in 2014. He is a Researcher with the Huawei Noah’Ark Lab, Shenzhen, China. His major interests include Time Series Analysis, Causality, and Deep learning. 
\end{IEEEbiography}

\vspace{-8ex}

\begin{IEEEbiography}
	[{\includegraphics[width=1in, height=1.25in, clip, keepaspectratio]{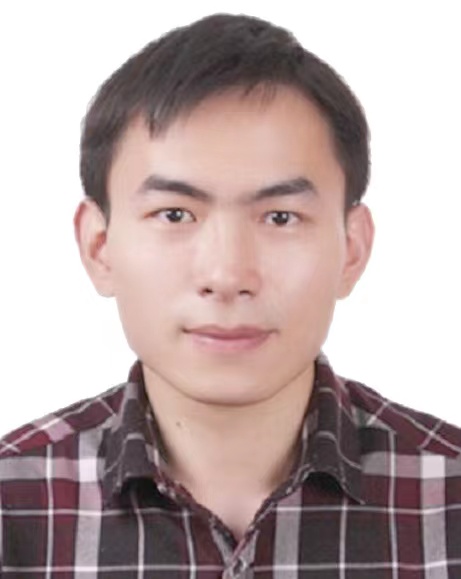}}]{Xi Zhang} received the MS.degree from Harbin Institute of Technology, Shenzhen, China in 2017. He is currently an application algorithm engineer in the department of assurance and managed services of Carrier Network Business Group. His main interest is to use AI technologies such as machine learning and deep learning to realize intelligent operation and maintenance of telecom networks.
\end{IEEEbiography}

\end{document}